\documentclass[11pt,onecolumn,fleqn]{IEEEtran}
                                                                                                        \usepackage{amsthm}
                                                                                                        \usepackage{amsfonts}
                                                                                                        \usepackage{amsmath}
                                                                                                        \usepackage{float}
                                                                                                        \usepackage{balance}
                                                                                                        \usepackage[ruled]{algorithm2e}
                                                                                                        \usepackage{verbatim}
                                                                                                        \usepackage[draft]{movie15}
                                                                                                        \usepackage{subfigure}
                                                                                                        \usepackage{graphicx}
                                                                                                       \usepackage{hyperref}

                                                                                                        \newfloat{Animation}{p}{ext}

                                                                                                        \newtheorem{definition}{Definition}
                                                                                                        \newtheorem{lemma}{Lemma}
                                                                                                        \newtheorem{corollary}{Corallary}
                                                                                                        \newtheorem{theorem}{Theorem}
                                                                                                        \newtheorem{claim}{Claim}

                                                                                                        \title{On the Workings of Genetic Algorithms\\The Genoclique Fixing Hypothesis}
                                                                                                        \author{Keki M. Burjorjee\\Computer Science Department\\Brandeis University\\Waltham, MA 02454\\ kekib@cs.brandeis.edu}

\begin{document}
                                                                                                        \maketitle
%

                                                                                                        \begin{abstract}
                                                                                                        We recently reported that the simple genetic algorithm (SGA) is capable of performing a remarkable form of sublinear computation which has a straightforward connection with the general problem of interacting attributes in data-mining. In this paper we explain how the SGA can leverage this computational proficiency to perform efficient adaptation on a broad class of fitness functions. Based on the relative ease with which a practical fitness function might belong to this broad class, we submit a new hypothesis about the workings of genetic algorithms. We explain why our hypothesis is superior to the building block hypothesis, and, by way of empirical validation, we present the results of an experiment in which the use of a simple mechanism called clamping dramatically improved the performance of an SGA with uniform crossover on large, randomly generated instances of the MAX 3-SAT problem.
                                                                                                        \end{abstract}

                                                                                                        \section{Introduction}
                                                                                                        Genetic algorithms are search heuristics that mimic natural evolution. They have been applied to a wide range of combinatorial optimization problems that are poorly understood, or known to be NP-Hard. While solutions generated by genetic algorithms are often inferior to those yielded by problem-specific search algorithms, in \emph{most} cases specialized search algorithms are not available. When used in such situations, genetic algorithms routinely generate usable solutions relatively quickly.

                                                                                                        Unfortunately, the workings of genetic algorithms (GAs) are not well understood. There are several anomalies in the empirical literature that cannot be explained by the \emph{building block hypothesis} \cite{Goldberg:1989:GAS,Holland92, Mitchell:1996:IGA}---the only comprehensive explanation for the adaptive capacity of genetic algorithms to be proffered to date. Of these anomalies, the two most serious are (i) the widely reported efficacy of uniform crossover \cite{Syswerda89, rudnick1994asg,pelikanSherrington}, and (ii) the unexpected behavior of GAs on  Royal Road functions  \cite{mitchell:1992:rrgaflgp, forrest93relative}.  In response to such anomalies, and to problems with the theoretical support for the building block hypothesis \cite{Fogel:1995:ECT,reeves2003gap}, the building block hypothesis is today treated with a certain amount of skepticism by many GA theorists.

                                                                                                        In distancing themselves from the building block hypothesis, several GA theorists have also moved away from the search for a single \emph{comprehensive} explanation for the adaptive capacity of genetic algorithms on practical problems, and have adopted what we shall call a \emph{many little theories} (MLT) approach. This approach is based on the belief that a single  theory about the practical workings of genetic algorithms is infeasible because genetic algorithms work in fundamentally different ways depending on, amongst other things, the operators they use, and the classes of practical optimization problems they are applied to. The goal of the MLT approach is to match classes of practical optimization problems with appropriate classes of genetic algorithms. By finding such matches, proponents of this approach hope, eventually, to supply GA practitioners with the means to determine the ``right" genetic algorithm for any practical problem.

                                                                                                        It seems unlikely that this vision will be realized anytime soon. For a small number of narrowly defined classes of fitness functions, researchers have had some success in deriving upper bounds on the expected number of fitness queries needed to find a global optimum (e.g. \cite{DBLP:journals/dam/JansenW05}). We are unaware, however, of any success in turning such theorems into \emph{theories}, even little ones, with demonstrable practical applications. Another dissatisfying feature of this approach is it's failure, to date, to identify a \emph{computational efficiency} of the genetic algorithm. i.e. a computation of some sort that the genetic algorithm can perform efficiently relative to other known algorithms. Most dissatisfying perhaps, especially to GA practitioners and would-be inventors of more powerful genetic algorithms, is the basic idea that a single comprehensive account of the practical workings of genetic algorithms is infeasible. A large section of the genetic algorithmics community seems to rejects this idea. Whether one accepts this idea or rejects it is a matter of one's metaphysics; we currently know of no definitive reason for deciding one way or the other. We should mention, however, that a viable comprehensive theory, if one can be found, is preferable, and that historically, scientists have been quite successful at finding viable comprehensive theories for large, internally diverse classes of systems. Most of those who reject the MLT idea continue to subscribe to some version of the building block hypothesis---weak theoretical foundation, and outstanding anomalies notwithstanding. The absence of a promising, \emph{comprehensive} alternative explains the entrenchment of this hypothesis. Presenting such an alternative is the aim of this paper.

                                                                                                        In a recent work \cite{DBLP:journals/corr/abs-0810-3357} we reported that the simple genetic algorithm (SGA) possesses a remarkable computational proficiency---a capacity for sublinear computation which, though irrelevant to the problem of global optimization, has straightforward connections with a currently intractable data-mining problem in computational genetics. In this paper, we demonstrate that by applying this computational proficiency recursively, an SGA can perform efficient adaptation on a specific class of fitness functions\footnote{We believe that the MLT community's inability to identify a computational efficiency of the SGA  is a consequence of it's strong focus on global optimization. This focus seems misplaced given that genetic algorithms are valued by practitioners, not for their capacity for efficient global optimization, but for their capacity for efficient \emph{adaptation}.}. Based on this result we infer that by recursively applying this computational proficiency, SGAs can perform efficient adaptation on a very broad class of fitness functions. Given the relative ease with which a practical fitness function might belong to this class of functions, we submit the genoclique fixing hypothesis---a new, comprehensive hypothesis about the practical workings of the simple genetic algorithm---and explain why, as comprehensive hypotheses go, this hypothesis is more promising than the building block hypothesis.

%
%
%

                                                                                                        If the genoclique fixing hypothesis is sound, it promises to precipitate significant improvements in the genetic algorithm's capacity for black-box combinatorial optimization. By way of empirical support for this hypothesis we describe what we consider to be the first of such improvements---a mechanism called \emph{clamping}---and present the results of an experiment in which the use of this simple mechanism dramatically improved the performance of a simple genetic algorithm with uniform crossover on large, randomly generated instances of the MAX 3-SAT problem \cite{hoos2004}.


%
                                                                                                        \subsection{Terminology}
                                                                                                        We use the word `gene' to refer to a genomic extent that tends not to be broken up by crossover. This usage accords with  Johansen's original use of this word, in 1909, to refer to a ``unit of inheritance" \cite{johansen09} \cite[p736]{mayr:gbt}. By this definition, a gene is not a strictly defined entity, but has a fading-out quality that is dependent on the expected number of crossover points, and the way these points tend to be distributed over a genome. There is no equivalent concept within genetic algorithmics. The notion of a building block \cite{Goldberg:1989:GAS} comes close, but since building blocks must, by definition, have above average fitness, whereas a gene need not, the two are not equivalent. It is important to stress that our use of the word gene differs from the way this word typical gets used in genetic algorithmics. Genetic algorithmicists tend to think of two adjacent genomic bits as two separate genes regardless of the crossover operator being used \cite{Mitchell:1996:IGA,Goldberg:1989:GAS}. We regard such bits as separate genes only when crossover is uniform, or close to uniform, i.e. when the expected number of crossover points is approximately half the value of the length of a genome. When the expected number of crossover points is significantly lower, these bits will tend to be inherited together. In this case we regard the two bits as two adjacent ``nucleotides" of a single gene.

                                                                                                        To ensure a clear comparison between our hypothesis and the building block hypothesis, we now express the latter using the terminology we have just adopted: The building block hypothesis assumes the existence in the initial population of large numbers of genes with  statistically significant fitness advantages. According to this hypothesis, adaptation in genetic algorithms is driven by the propagation of such genes, and by the frequent composition in offspring of co-adapted sets of individually advantageous genes that are not co-present in either parent. To avoid confusion, it is important to clarify that by `co-adapted' we mean something other than the existence of super-additive, or super-multiplicative fitness interactions between the the genes concerned; rather, we mean simply that the expected fitness of a genome carrying all the genes in the `co-adapted' set is greater than the expected fitness of a genome carrying any individual gene in the set; the whole, in other words, is greater than any of the parts.

                                                                                                        \subsection{The Basic Idea}
                                                                                                        We have previously reported \cite{DBLP:journals/corr/abs-0810-3357} that an SGA is capable of efficiently driving a set of co-adapted, \emph{unlinked} genes to fixation even though the fitness signal of this set of genes may be weak relative to the background noise. In driving such genes to fixation the SGA raised the average fitness of the population by a small amount.  When any set of genes gets fixed in the population, the representation of the problem space can be thought to have changed. Crucially, the new representation may contain one or more sets of co-adapted genes which may not have had a detectable fitness signal in the old representation. By subsequently driving one or more of these sets to fixation, the SGA can once again ``change" it's representation, and in doing so can create new small sets of coadapted genes. And so on.

                                                                                                        Each time a small set of co-adapted genes gets fixed, the average fitness of the population will increase by an amount that may be tiny. As the fixation of small sets of co-adapted genes continues, however, these amounts will begin to add up. Based on this thought experiment, we hypothesize that adaptation in genetic algorithms is driven by the iterated  ``creative fixation" of small sets of co-adapted genes.

                                                                                                        \section{The Genoclique Fixing Hypothesis}

                                                                                                        Our hypothesis pertains to the class of recombinative SGAs. Our model for this class is the simple genetic algorithm with uniform crossover (UGA). We adopt this algorithm as our model for two reasons: Firstly, under uniform crossover the notion of a unit of inheritance, i.e. a gene, is crisply defined---a gene corresponds exactly with a single bit in a bitstring. This conceptual crispness greatly simplifies our exposition. Secondly, by using suitably crafted classes of fitness functions, the absence of positional bias \cite{eshelman1989bcl} in uniform crossover can be exploited to demonstrate the computational efficiencies that form the basis for our hypothesis.

                                                                                                        \subsection{Mathematical Preliminaries}
                                                                                                        For any positive integer $\ell$, we denote the set of all bitstrings of length $\ell$ by $\mathfrak B_\ell$. We denote a \emph{schema partition} \cite{Mitchell:1996:IGA} by a tuple consisting of the indices of the defining positions of that schema partition---e.g. $(2, 15, 3)$. The \emph{order} of a schema partition $\Gamma$, denoted by $o(\Gamma)$, is the number of elements in some tuple that denotes $\Gamma$. Note that a tuple that denotes some schema partition does not have to be ordered; therefore, schema partitions with order greater than one can be denoted in more than one ways.  Let $\Gamma_1$ and $\Gamma_2$ denote two schema partitions. We say that these schema partitions are \emph{orthogonal} if the tuples $\Gamma_1$ and $\Gamma_2$ have no elements in common. For any genome $g$, let $g_i$ denote the $i^{th}$ bit of $g$. For any positive integer $n$, let $[n]$ denote the set $\{1,\ldots,n\}$. For any genome $g$ of length $\ell$ and any $k$-tuple $x$ of distinct integers in $[\ell]$, let $\Xi_x(g)$ denote the bitstring $g_{x_1}\ldots g_{x_k}$. The denotation of a \emph{schema} is dependent on the denotation of the schema-partition that the schema belongs to. Given a schema partition denoted by some tuple $\Gamma$, the schemata in this partition are denoted by bitstrings of length  $o(\Gamma)$.  For any bits $b_1,\ldots,b_{o(\Gamma)}$, the bitstring $b_1\ldots b_{o(\Gamma)}$ denotes the schema consisting of the genomes $ \{g|\Xi_{\Gamma}(g)=b_1\ldots b_{o(\Gamma)}\}$. The denotation of the relevant schema partition must always be borne in mind when interpreting a denoted schema.

                                                                                                        Let $\Gamma_1=(x_1, \ldots, x_m)$ and $\Gamma_2=(y_1,\ldots,y_n)$ denote two orthogonal schema partitions, and let $\gamma_1=a_1\ldots a_n$ and $\gamma_2=b_1\ldots b_n$ denote schemata of  $\Gamma_1$ and $\Gamma_2$ respectively. Then the concatenation $\Gamma_1\Gamma_2$ denotes the schema partition $(x_1, \ldots x_m,y_1, \ldots, y_n)$, and the concatenation $\gamma_1\gamma_2$ denotes the schema $a_1\ldots a_mb_1\ldots b_n$ of $\Gamma_1\Gamma_2$. We will treat the denotation of a schema partition as a tuple sometimes, and as the represented schema partition at others. Likewise, we will treat the denotation of a schema as a bitstring sometimes, and as the represented schema at others. The sense in which we use the denotations of schemata and schema partitions will be clear from the context. For any $m\times n$ matrix $M$, and for any $i\in[m]$, let $M_{i:}$ denote the $n$-tuple that the row $i^{th}$ row of  $M$.

                                                                                                        \subsection{Staircase Functions}

                                                                                                        We begin by introducing a class of fitness functions such that the co-adaptedness of most small sets of bits---genes, if we assume that crossover is uniform---is highly contingent upon the fixation of other bits.

                                                                                                        \begin{definition} A staircase function descriptor is a 7-tuple $(h,o,\delta,\sigma, \ell, L,V)$ where $h$, $o$ and $\ell$ are positive integers with $ho\geq\ell$, $\delta$ and $\sigma$ are positive real numbers, and $L$ and $V$ are matrices with $h$ rows and $o$ columns such that the values of $V$ are binary digits, the elements of  $L$ are distinct integers from the set $[\ell]$, and the rows of $L$ are sorted in ascending order. \end{definition}

                                                                                                        Let $\mathcal N(a,b)$ denote the normal distribution with mean $a$ and variance $b$. Then the function described by a staircase function descriptor $(h, o, \delta,\sigma, \ell, L, V)$ is the stochastic function over the set of bitstrings of length $\ell$ given by algorithm 1. We call $h, o, \delta$, $\sigma$, and $\ell$ the \emph{height, order, increment, noisiness} and \emph{span}, respectively, of the staircase function.

                                                                                                             \begin{algorithm}[tb!]
                                                                                                                   \dontprintsemicolon
                                                                                                                   \KwIn{$g$ is a genome of length $\ell$}\;
                                                                                                                   $y=$\mbox{some value drawn from the distribution $\mathcal N(0,\sigma^2)$}\;
                                                                                                                   \For{i=1 to h}{
                                                                                                                         \eIf{$\Xi_{L_{i:}}(g)=V_{i1}\ldots V_{io}$}
                                                                                                                            {$y=y+\delta$}
                                                                                                                            {$y=y-(\delta/(2^o-1))$\;
                                                                                                                               \textbf{break}
                                                                                                                         }
                                                                                                                   }
                                                                                                                   \KwRet{$y$}\;
                                                                                                                   \caption[boo]{\\A staircase function with descriptor $(h,o,\delta,\sigma,\ell,L,V)$}
                                                                                                             \end{algorithm}

                                                                                                        For any $i \in [h]$ we call the schema denoted by $V_{i1}\ldots V_{io}$ of the schema partition denoted by $(L_{i1},\ldots,L_{io})$ the $i^{th}$ \emph{stage} of the staircase function $f$. Given the matrix $L$ of the staircase function descriptor, the schema partition of each stage has a canonical denotation. When the staircase function descriptor is clear we will, in the interest of concision, assume that the schema partition of each stage is denoted canonically. Let $\gamma_i$ denote the $i^{th}$ stage of $f$. We call the schema denoted by $\gamma_1\ldots\gamma_i$ the $i^{th}$ \emph{step} of $f$.

                                                                                                        The steps of a staircase function are essentially a progression of nested \emph{hyperplanes} \cite[p 53]{Goldberg:1989:GAS}, with hyperplanes of higher order and higher expected fitness nested within hyperplanes of lower order and lower expected fitness. By choosing an appropriate scheme for mapping a high-dimensional hypercube onto a two dimensional plot, it becomes possible to visualize this progression of hyperplanes in two dimensions.

                                                                                                        \begin{algorithm}[b!]
                                                                                                        \dontprintsemicolon
                                                                                                        \KwIn{$g$ is a genome of length $2mn$}\;
                                                                                                        $granularity=2^{mn}/2^n$\;
                                                                                                        $x=0$\;
                                                                                                        $y=0$\;
                                                                                                        \For{$i=1$ to $m$}{
                                                                                                            $x=x+granularity*bin2Int (\Xi_{X_{i:}}(g))$\;
                                                                                                            $y=y+granularity*bin2Int (\Xi_{Y_{i:}}(g))$\;
                                                                                                            $granularity=granularity/2^n$\;
                                                                                                        }
                                                                                                        \KwRet{$x, y$}\;
                                                                                                        \caption{\label{fractaladdressing}The algorithm for determining the ($x$, $y$)-address of a genome under the fractal addressing system $(m,n,X,Y)$. The function $bin2Int$ returns the integer value of a binary string.}
                                                                                                        \end{algorithm}

                                                                                                        \begin{definition}
                                                                                                        A fractal addressing system is a tuple $(m,n, X,Y)$, where $m$ and $n$ are positive integers, and $X$ and $Y$ are matrices with $m$ rows and $n$ columns such that the elements in $X$ and $Y$ are distinct positive integers from the set $[2mn]$, i.e. each element in $[2mn]$ occurs either in $X$ or in $Y$ once and only once.
                                                                                                        \end{definition}

                                                                                                        A fractal addressing system $(m,o,X,Y)$  determines how the set $\mathfrak B_{2mn}$ gets mapped onto a $2^{mn}\times 2^{mn}$ plot. For any bitstring $g\in\mathfrak B_{2mn}$ the $xy$-address (a tuple of values between 1 and $2^{mn}$) of the pixel representing $g$ is given by Algorithm \ref{fractaladdressing}.

\begin{figure*}[tb!]\begin{center}
                                                                                                        \subfigure[]{\includegraphics[width=.4\textwidth]{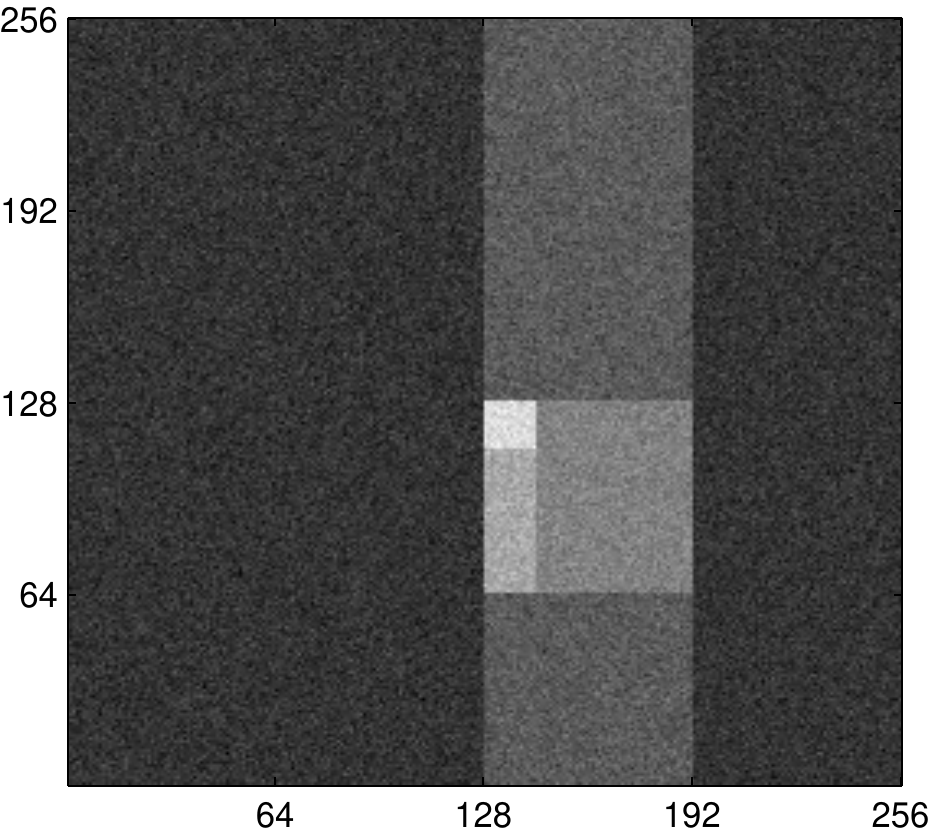}}\quad\quad\quad\quad
                                                                                                        \subfigure[]{\includegraphics[width=.4\textwidth]{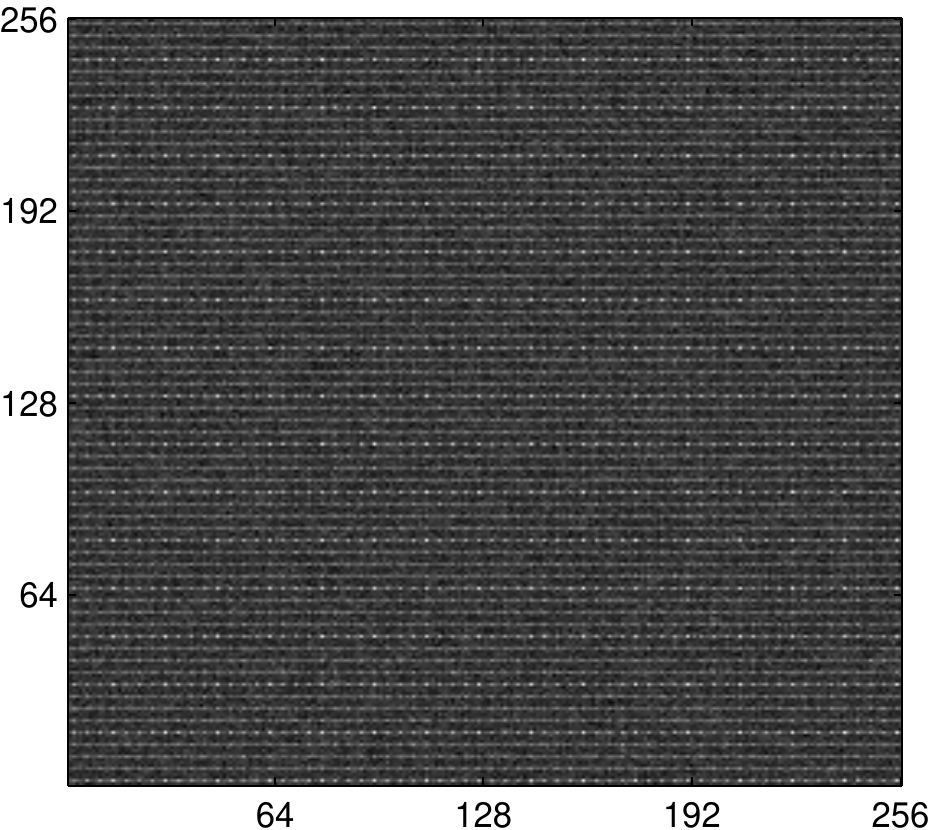}}\end{center}\caption{\label{ladderfunctionvis} A fractal plot of the staircase function $f$ under the fractal addressing systems $A$ (\emph{left}) and $A'$ (\emph{right})}
\end{figure*}

                                                                                                        \noindent\textbf{Example: } Let $(h=4,o=2,\delta=3,\sigma=1,\ell=16,L,V )$ be the descriptor of a basic pivotal function $f$, such that \[V=\left[\begin{array}{cc}1&0\\0&1\\0&0\\1&1\end{array}\right]\] Let $A=(m=4,n=2,X,Y)$ be a fractal addressing system such that  $X_{1:}=L_{1:}$,  $Y_{1:}=L_{2:}$, $X_{2:}=L_{3:}$, and $Y_{2:}=L_{4:}$. A \emph{fractal plot} of $f$ is shown in Figure \ref{ladderfunctionvis}a.

                                                                                                        This image was generated by querying $f$ with every bitstring in $\mathfrak B_{16}$, and plotting the resulting fitness value of each genome as a greyscale pixel at the genome's fractal address (under the addressing system $A$). The fitness values returned by $f$ have been scaled linearly to span the range of possible greyscale shades. Lighter shades signify greater fitness. The four steps of $f$ can easily be discerned.

                                                                                                        Let us perform a thought experiment in which we generate another fractal plot of $f$ using the same addressing system $A$, but a different random number generator seed. Because $f$ is stochastic, the greyscale value of any pixel in the resulting plot will then most likely be different from that of its homolog in the the plot in Figure \ref{ladderfunctionvis}a. Nevertheless, our ability to discern the steps of $f$ would not be affected. In the same vein, note that when specifying $A$, we have not specified the values of the last two rows of $X$ and $Y$; it is easily seen that these values are immaterial to the discernment of the staircase structure of $f$.

                                                                                                        On the other hand, the values of the first two rows of $X$ and $Y$ are highly relevant to the discernment of this structure.  Figure \ref{ladderfunctionvis}b shows a fractal plot of $f$ that was obtained using a fractal addressing system  $A'=(m=4,n=2, X',Y')$ such that $X'_{4:}=L_{1:}$,  $Y'_{4:}=L_{2:}$, $X'_{3:}=L_{3:}$, and $Y'_{3:}=L_{4:}$. Nothing remotely resembling a staircase is visible in this plot.

                                                                                                        The lesson here is that the discernment of the fitness staircase inherent within a staircase function depends critically on how one `looks' at this function. In determining the `right' way to look at $f$ we have used information about the descriptor of $f$, specifically the values of, $h,o$, and  $L$. This information will not be available to an algorithm which only has query access to $f$.

\begin{figure*}[tb!]
                                                                                                        \begin{center}\subfigure{\includegraphics[width=.4\textwidth]{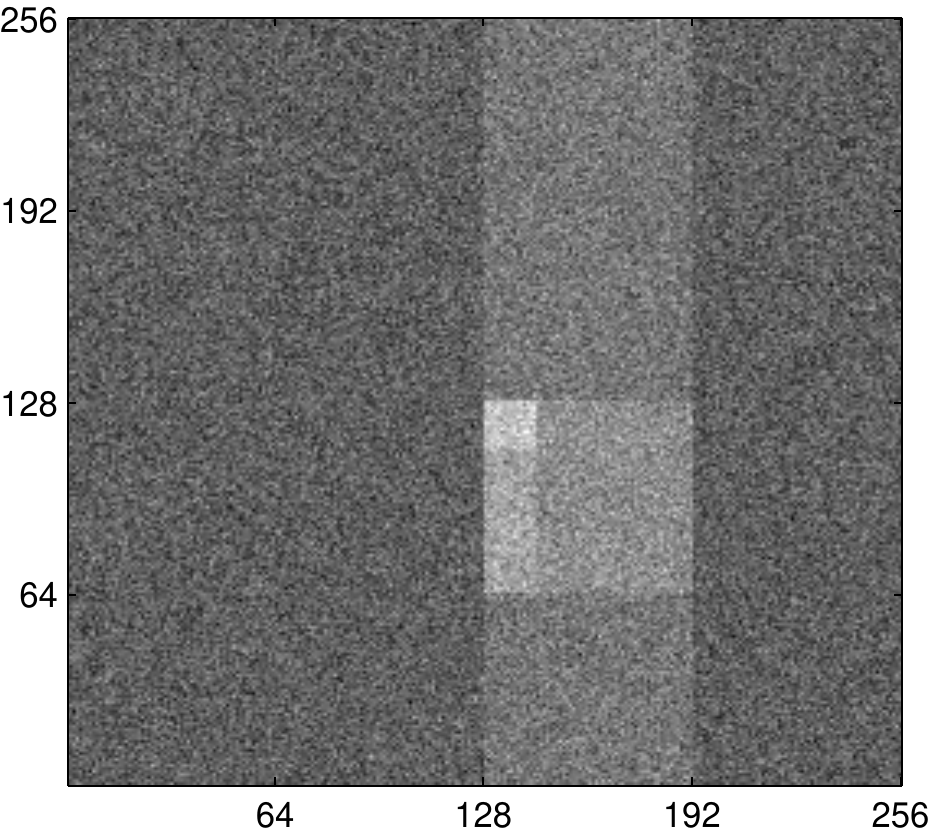}}
                                                                                                        \quad\quad\quad\subfigure{\includegraphics[width=.4\textwidth]{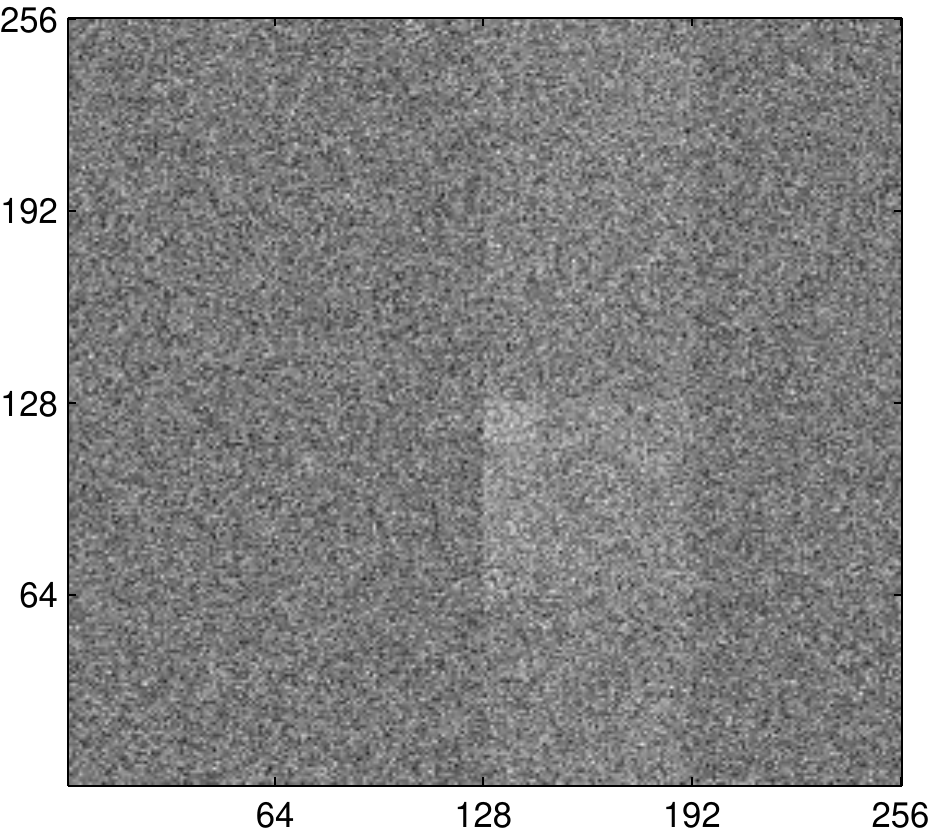}}\end{center}
                                                                                                        \caption{\label{fractalPlot2} Fractal plots under $A$ of two staircase functions, which differ from  $f$ only in their increments---1 \emph{(left plot)} and 0.3 \emph{(right plot)}  as opposed to  3 }.
\end{figure*}

                                                                                                        Even if one knows the right way to look at a staircase function, the discernment of the fitness staircase inherent within this function can still be made difficult by a low increment to noisiness ratio. Figure \ref{fractalPlot2} lets us visualize the decrease in the salience of the fitness staircase of $f$ that accompanies a decrease in the increment parameter of this staircase function. As mentioned before, the fitness values returned by the staircase functions are scaled so that they span the range of possible greyscale shades; therefore, had we kept the increment constant and increased the noisiness parameter instead, we would have obtained the same general result as that shown in Figure \ref{fractalPlot2}. In general, a decrease in the increment to noisiness ratio of a staircase function results in a decrease in the `contrast' between the steps of that function.

                                                                                                         Let $\gamma$ denote some schema of the schema partition denoted by $\Gamma$. Given some (possibly stochastic) fitness function over a genome set, we define the \emph{fitness signal} of $\gamma$, denoted $S_{\Gamma}(\gamma)$  to be the expected fitness of a genome drawn from the uniform distribution over $\gamma$. Let $\gamma_1$ and $\gamma_2$ be schemata of two orthogonal schema partitions $\Gamma_1$ and $\Gamma_2$.  We define the \emph{conditional fitness signal of}  $\gamma_1$ \emph{given} $\gamma_2$, denoted $S_{\Gamma_1|\Gamma_2}(\gamma_1\,|\,\gamma_2)$, to be the difference between the fitness signal of  $\gamma_1\gamma_2$ and the fitness signal of $\gamma_2$, i.e. $S_{\Gamma_1|\Gamma_2}(\gamma_1\,|\,\gamma_2)=S_{\Gamma_1\Gamma_2}(\gamma_1\gamma_2)-S_{\Gamma_2}(\gamma_2)$.

                                                                                                         Given a staircase function $f$ with descriptor $(h,o,\delta,\sigma,\ell, L,V)$, we define the \emph{signal to noise ratio} of some schema $\gamma$ of a schema partition $\Gamma$ to be $S_\Gamma(\gamma)/\sigma$. Likewise, for any two schemata $\gamma_1$ and $\gamma_2$ of two orthogonal schema partitions $\Gamma_1$ and $\Gamma_2$, we define the \emph{conditional signal to noise ratio} of $\gamma_1$ given $\gamma_2$ to be  $S_{\Gamma_1|\Gamma_2}(\gamma_1\, |\,\gamma_2)/\sigma$.

                                                                                                         For any $i\in[h]$, by Lemma 1 (see appendix), the signal to noise ratio of step $i$ is $i\delta/\sigma$. For any $i\in\{2,\ldots,h\}$, corollary 1 of Lemma 1 states that the  \emph{conditional} signal to noise ratio of stage $i$ \emph{given} step $(i-1)$ is  $\delta/\sigma$, (a \emph{constant} with respect to $i$). Finally, for any $i\in[h]$, by Theorem 1, the (unconditional) signal to noise ratio of stage $i$ is \begin{equation}\frac{\delta}{(2^o)^{i-1}\sigma}\end{equation} Clearly, this ratio decreases rapidly as $i$ increases.

                                                                                                           Consider an algorithm that, when given only query access to the staircase function $f$, can robustly detect the fitness signal of  the first step of $f$, and can restrict future sampling to this step. Observe that the \emph{conditional } signal to noise ratio of the second stage  \emph{given} the first step is the same as the signal to noise ratio of  the first step. Therefore, if the algorithm restricts its fitness queries to genomes belonging to the first step, it should be able to detect the \emph{conditional} fitness signal of the second stage \emph{given} the first step, and should, therefore, be able to identify the second step.  Indeed if the algorithm is sufficiently robust it's recursive application need not end with the identification of the second step; higher steps can be identified \emph{indirectly} by identifying lower steps first.

                                                                                                         Given expression (1), it is reasonable to suspect that the \emph{direct} identification of step $i$ of a staircase function rapidly becomes computationally infeasible as $i$ increases. The analogy between physical staircases and staircase functions should be transparent; just as it is hard to climb higher steps of a staircase without climbing lower steps first, it becomes computationally infeasible to identify higher steps of a staircase function without identifying lower steps first.

                                                                                                         \subsection{\label{hyperclimbing}Hyperclimbing and Hyperscapes}

                                                                                                         When an algorithm restricts future queries to some step of a staircase function, we say that it has \emph{climbed} that step. The idea of climbing the steps of a staircase function can be generalized to describe the behavior of arbitrary search algorithms on arbitrary fitness functions (both stochastic and deterministic) over sets of strings. We call the progressive confinement of sampling to hyperplanes of increasing order and increasing expected fitness \emph{hyperclimbing} (short for ``hyperplane-climbing"); a search algorithm is said to have \emph{climbed} some hyperplane $\gamma$ that belongs to some hyperplane partition $\Gamma$, if, amongst all the hyperplanes that belong to $\Gamma$, future sampling is largely limited to the hyperplane $\gamma$.

                                                                                                         Hyperclimbing, if it can be implemented efficiently (a big \emph{if}), seems like a very reasonable way to perform adaptation. Consider some practical fitness function over the set of bitstrings $\mathfrak B_\ell$. It is seems reasonable to assume that there exists some low number $o_1$, such that of the ${\ell_{\phantom 1} \choose o_1}\in\Omega(\ell^{o_1})$ ways of partitioning the search space into a set of hyper planes of order $o_1$, there exists one or more partitions---for the sake of argument let us assume just one---such that this partition contains one or more hyperplanes whose average fitness values are statistically significantly above average under uniform sampling. By restricting future sampling to one of these hyperplanes the hyperclimbing heuristic can increase the expected fitness of all future samples. As far as the hyperclimbing heuristic is concerned, this hyperplane would then  comprise the entirety of the search space, i.e. future search can be thought to occur over the space $\mathfrak  B_{\ell-o_1}$. Our argument now recurses:  It seems reasonable to assume that there exists some low number $o_2$, such that of the ${\ell-o_1 \choose o_2}\in\Omega((\ell-o_1)^{o_2})$ ways of partitioning the new search space into a set of hyperplanes of order $o_2$, there exists one or more partitions---for the sake of argument let us assume just one---such that this partition contains one or more hyperplanes whose average fitness values are statistically significantly above average under uniform sampling. By restricting future sampling to one of these hyperplanes, the hyperclimbing heuristic would, once again, increase the expected fitness of all future samples. And so on.

                                                                                                         This heuristic will continue to increase the average fitness of the samples it generates as long as there continues to be a way of partitioning the region of the the search space that it inhabits into a set of low-order hyperplanes such that at least one hyperplane in the partition has an average fitness value that is statistically significantly above average under uniform sampling.

                                                                                                          Because a hyperclimbing heuristic is sensitive to the ``hyperplanar structure" of a search space, not its neighborhood structure, the idea of a landscape \cite{WrightLandscape,kauffman1993oos} is not very helpful when thinking about the behavior of this heuristic. Far more useful is the notion of a \emph{hyperscape}. A hyperscape is like a landscape in that it is just a spatial representation of a fitness function. In a hyperscape, however, the focus is placed, not on the interplay between the fitness function and the neighborhood structure of individual points, but on the statistical fitness properties of individual hyperplanes,  and on the spatial relationships between hyperplanes---lower order hyperplanes can \emph{contain} higher order hyperplanes, hyperplanes can \emph{intersect} each other, and disjoint hyperplanes that belong to the same hyperplane partition can be regarded as \emph{parallel}. The use of the concept of a hyperscape in the genetic algorithmics literature can be traced back to the seminal work of Holland \cite{holland75:_adapt_natur_artif_system}, who used this concept to reason about the dynamics of recombinative genetic systems. While we disagree with Holland's conclusions, we find hyperscapes to be invaluable in our own reasoning about the dynamics of genetic algorithms---both recombinative and, for reasons that will become clear in section \ref{functionOfRecombination}, non-recombinative.

                                                                                                          \subsection{Symmetry Analysis}

                                                                                                          In a recent work \cite{DBLP:journals/corr/abs-0810-3357} we defined the class of \emph{semi-parameterized} UGAs, and exploited the symmetries of the algorithms in this class to uncover what we consider to be the first two computational efficiencies (albeit highly specific ones) of the SGA to be rigorously identified. The symmetry analysis in that work sets the stage for the symmetry analysis given below. We will show that a semi-parameterized UGA can efficiently climb the first few steps of the staircase functions in a particular class of staircase functions. Remarkably the number of queries required by the semi-parameterized UGA is \emph{independent} of the span of the functions in the class.

                                                                                                          Let $f$ be a staircase function with descriptor $(h,o,\delta,\sigma, \ell, L, V)$, we say that this function is \emph{basic} if $\ell=ho$, $L_{ij}=o(i-1)+j$, (i.e. if $L$ is the matrix of integers from 1 to $ho$ laid out row-wise), and $V$ is a matrix of ones. If $f$ is basic, then the last three elements of the descriptor of $f$ are fully determinable from the first four; we therefore write this descriptor as $(h,o,\delta,\sigma)$. Given some staircase function $f$ with descriptor $(h,o,\delta,\sigma, \ell, L, V)$, we define the \emph{basic form} of $f$ to be the (basic) staircase function with descriptor $(h,o,\delta,\sigma)$.

                                                                                                         Let $f^*$ be some basic staircase function with descriptor $(h,o,\delta,\sigma)$, and let $F$ be the set of all staircase functions with basic form $f^*$. Let $W$ be a semi-parameterized UGA. For any staircase function $f\in F$, let  $p^{(t)}_{(f,i)}(x)$ be the probability that the frequency of \emph{stage} $i$ of $f$ in generation $t$ of $W^{f}$ is $x$, let $q^{(t)}_{(f,i)}(x)$ be the probability that the frequency of \emph{step} $i$ of $f$ in generation $t$ of $W^{f}$ is $x$, and let $r^{(t)}_f$ be the probability that the average fitness of the population of $W^{f}$ in generation $t$ is $x$. Then by appreciating the symmetries between the unparameterized UGAs $W^{f^*}$ and $W^{f}$ we can deduce the following equalities between probability distributions: for any generation $t$, and for any $i\in[h]$, $p^{(t)}_{(f,i)}=p^{(t)}_{(f^*,i)}$, $q^{(t)}_{(f,i)}=q^{(t)}_{(f^*,i)}$, and $r^{(t)}_{f}=r^{(t)}_{f^*}$.

                                                                                                         Thus, for any generation $t$, monte-carlo sampling from $r^{(t)}_{f^*}$  is equivalent to monte-carlo sampling from $r^{(t)}_{f}$, and for any $i\in[h]$, monte-carlo sampling from $p^{(t)}_{(f^*,i)}$, and $q^{(t)}_{(f^*,i)}$ is equivalent to monte-carlo sampling from $p^{(t)}_{(f,i)}$, and $q^{(t)}_{(f,i)}$ respectively.

                                                                                                          \subsection{Performance of UGAs on a Staircase Function}

\begin{figure*}[tb!]\begin{center}
                                                                                                         \subfigure[Performance of the UGA $U^{f_1}$]{\includegraphics[width=7cm]{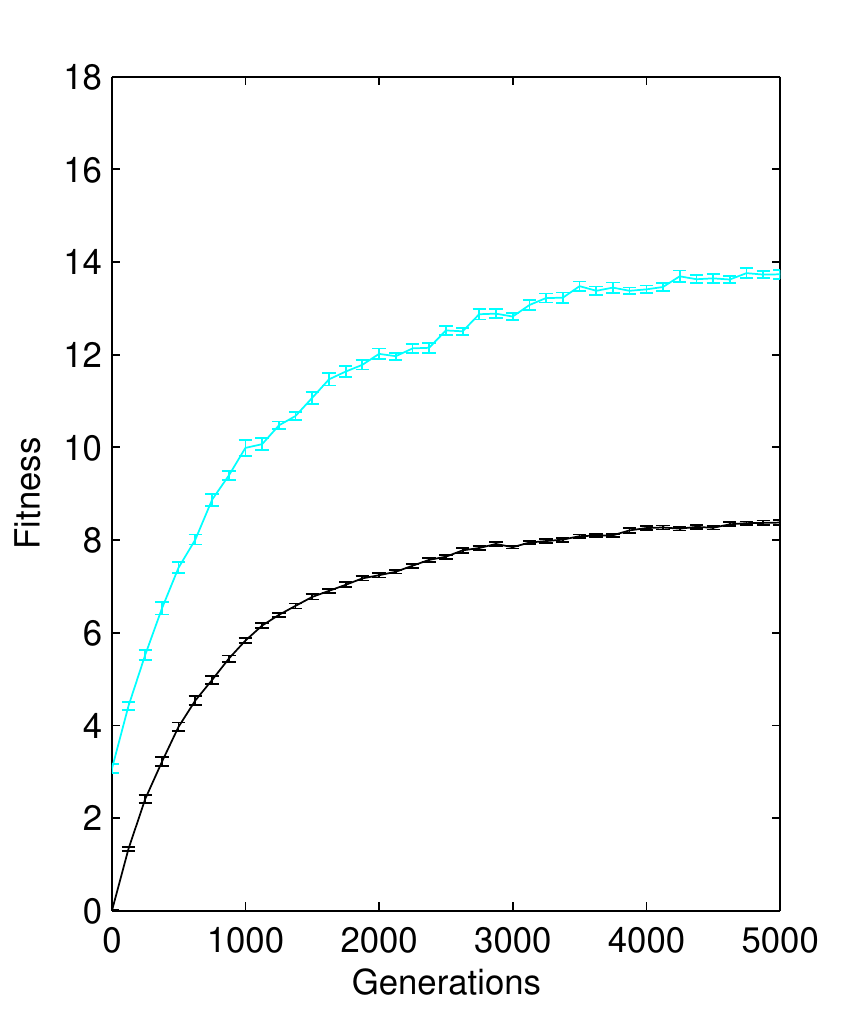}}\quad\quad\quad
                                                                                                         \subfigure[Performance of the MGA $M^{f_1}$]{\includegraphics[width=7cm]{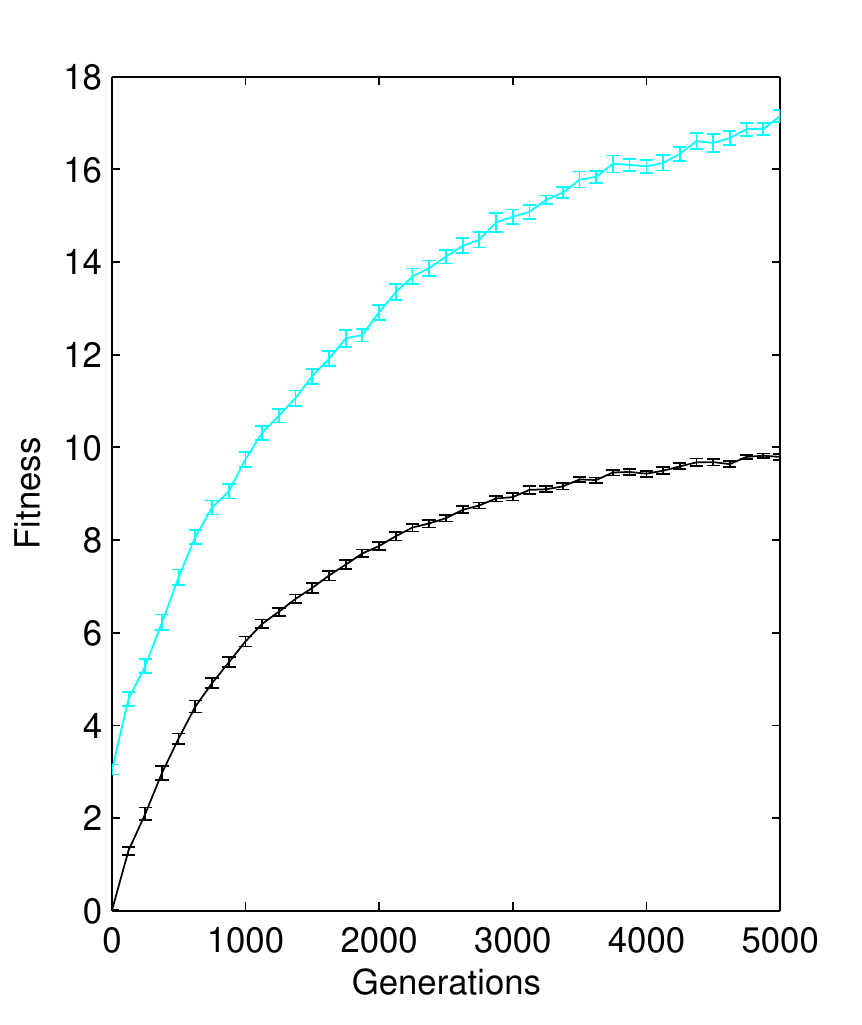}}\end{center}
                                                                                                         \caption{\label{crosstype=0Performance}The performance of the semi-parameterized UGA $U$ (left) and the semi-parameterized MGA $M$ (right) on the staircase function $f_1$ over 20 trials. The mean (across trials) of the average fitness of the population is shown in black. The mean of the best-of-population fitness is shown in blue. The error bars show one standard error above and below the mean every $125^{th}$ generation}
\end{figure*}

\begin{figure*}[tb!]\begin{center}
                                                                                                         \subfigure[Frequencies of first seven steps of $f_1$ under the action of $U$]{\includegraphics[width=7cm]{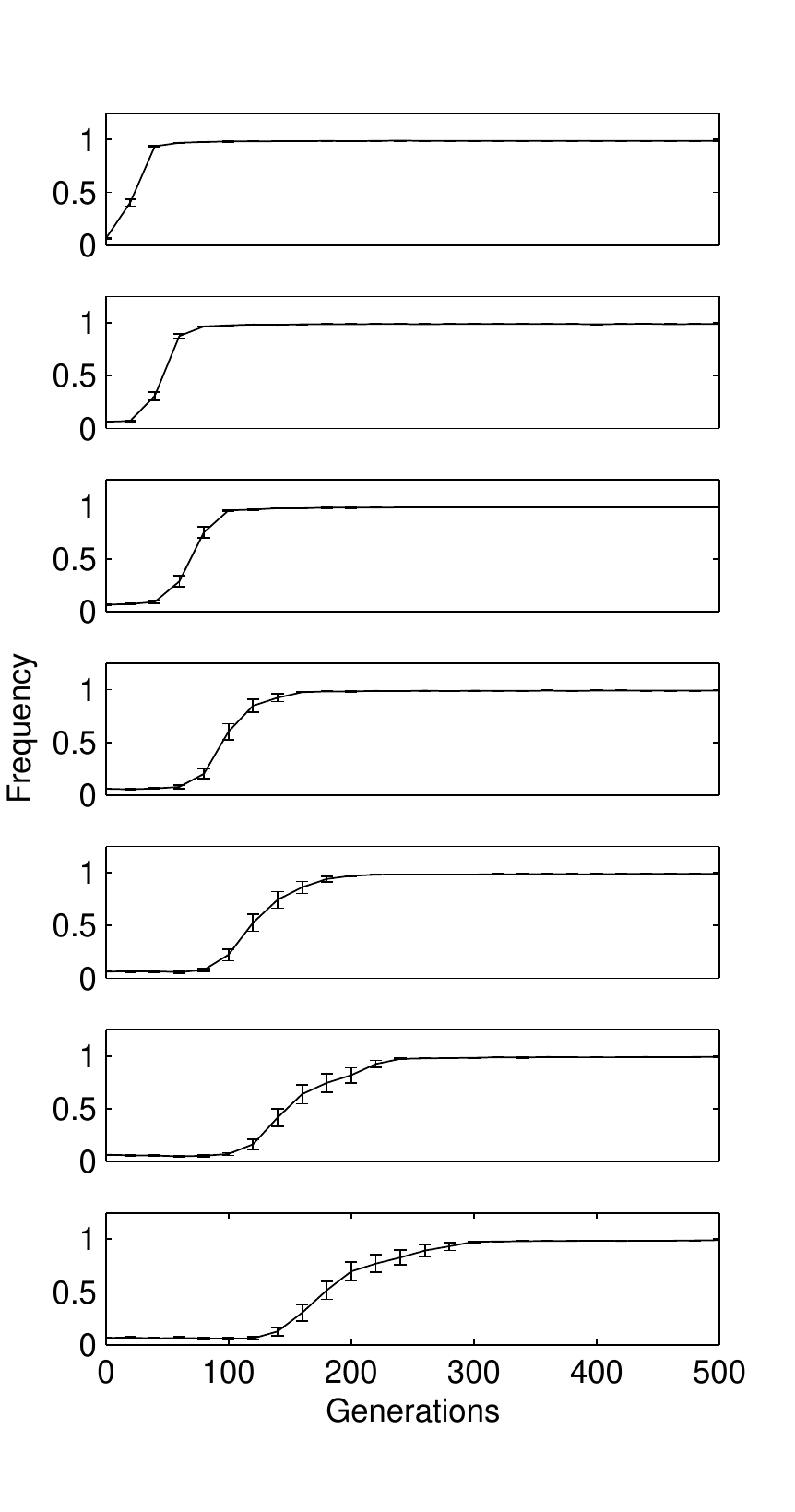}}\quad\quad\quad
                                                                                                         \subfigure[Frequencies of first seven steps of $f_1$ under the action of $M$]{\includegraphics[width=7cm]{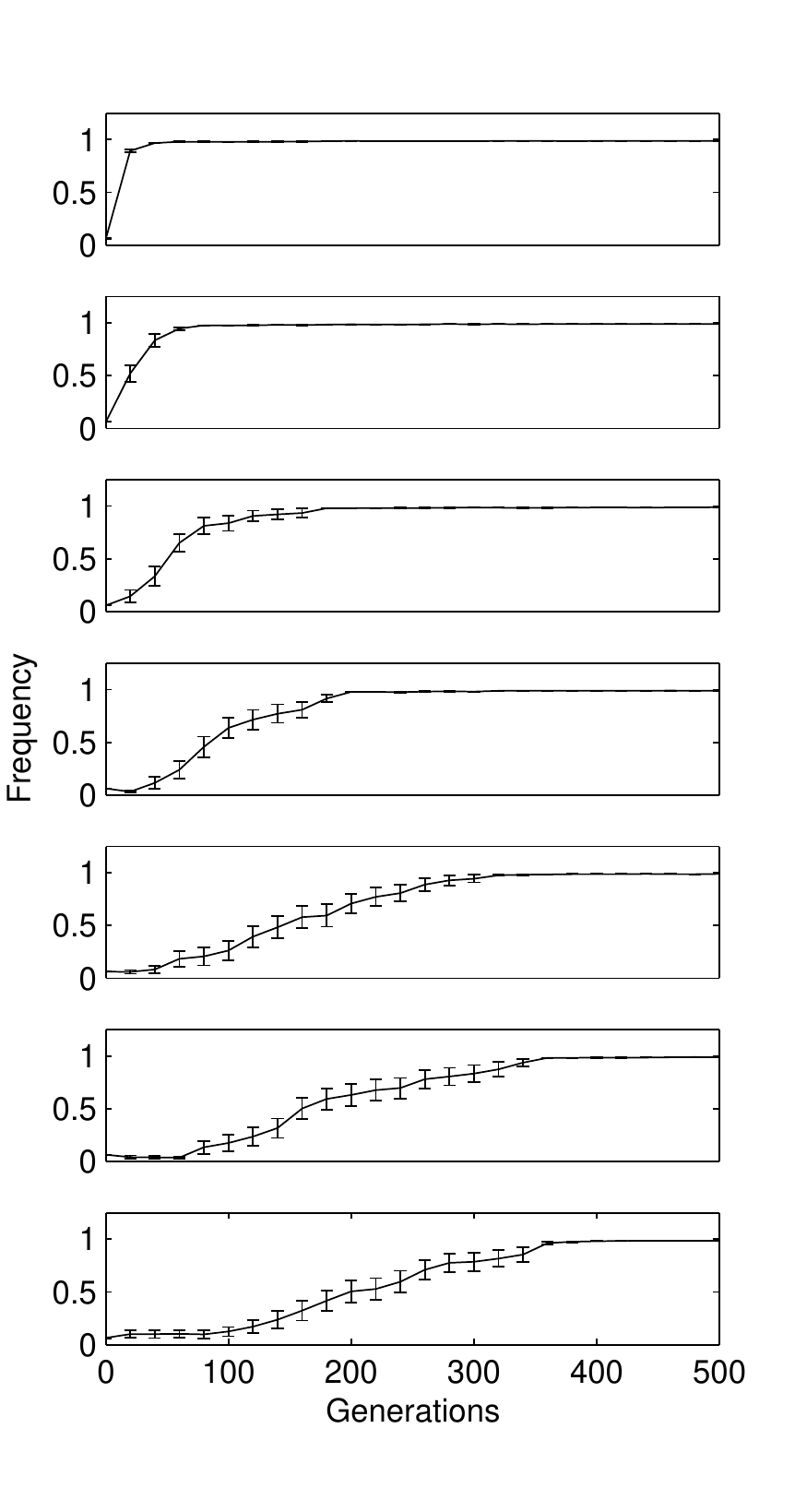}}\end{center}
                                                                                                         \caption{\label{crosstype0Frequencies}The mean frequency dynamics, over 20 trials, of the first seven steps of the staircase function $f_1$ (going from the top plot to the bottom plot) under the action of the semi-parameterized UGA $U$ (left), and the semi-parameterized MGA $M$ (right).  The error bars show one standard error above and below the mean every twentieth generation}

\end{figure*}

                                                                                                         Let $f_1$ be a staircase function with descriptor $(h=50, o=4, \delta=0.3, \sigma=1)$, and let $U$ denote the semi-parameterized UGA described in the materials and methods section
                                                                                                         in the appendix. In order to succinctly discuss the results of an experiment in which we applied $U$ to $f_1$, we introduce the following shorthand: given some population of genomes, the \emph{one-frequency} of some locus is the frequency of the bit 1 at that locus in the population. Figure \ref{crosstype=0Performance}a shows that $U$  is capable of robust adaptation when applied to $f_1$. Figure \ref{crosstype0Frequencies}a shows that under the action of $U$, the first seven stages of $f_1$ tend to go to fixation\footnote{We use the terms `fixation' and `fixing' loosely. Clearly, as long as the mutation rate is non-zero, no locus can ever be said to go to fixation in the strict sense of the word.}
                                                                                                          in ascending order. This entails that the first seven steps tend to go to fixation in ascending order. When a step gets fixed, future sampling will largely be confined to that step---in effect, the hyperplane associated with the step has been climbed. Animation \ref{crosstype2Mut003Anim}, which plots the one-frequencies of all the loci of $U^{f_1}$ in each of 500 generations, shows that the hyperclimbing behavior of $U^{f_1}$ continues beyond the first seven steps. The capacity of  $U$ to implement hyperclimbing when applied to $f_1$ accounts for it's adaptive ability on $f_1$.

                                                                                                         Let $f$ be some staircase function with basic form $f_1$. The conclusions reached in the previous section entail that, had we applied $U$ to $f$ instead of $f_1$, then \emph{regardless of the span of} $f$,   we would have obtained essentially the same results as those shown in Figures \ref{crosstype=0Performance}a and \ref{crosstype0Frequencies}a. This realization is highly remarkable from a computational standpoint.

                                                                                                         Consider  $U$'s capacity for  climbing just the first stage of $f$.  From a computational standpoint, even just this ability is quite remarkable because it is achieved with an expected expenditure of queries that is \emph{constant} in the span of $f$.  We infer that this highly specific capacity for computational efficiency is part of a general capacity of the SGA for efficiently performing what we call \emph{genoclique fixing}. We have previously identified two other highly specific, but nonetheless remarkable, computational efficiencies of the SGA that are instances of  its general capacity for efficient genoclique fixing \cite{DBLP:journals/corr/abs-0810-3357}. The results presented here suggest that SGAs can engender robust and efficient adaptation by performing efficient genoclique fixing \emph{recursively}.

                                                                                                         \subsection{Mutational Drag and Clamping}

                                                                                                         Before discussing genoclique fixing, let us contemplate a curious aspect of the behavior of $U$ on $f_1$. Figure 1 shows that the growth rate of the average fitness of the population of $U^{f_1}$ decreases as evolution proceeds. To understand this phenomenon consider some genome that belongs to the $i^{th}$ step; the probability that this genome will still belong to step $i$ after mutation is $(1-x) ^{io}$, where $x$ is the per-bit mutation rate. This entails that,  $U^{f_1}$ becomes less able to ``hold" a population within step $i$ as $i$ increases. In light of this observation, we infer that as $i$ increases the capacity of $U^{f_1}$ to be sensitive to the conditional fitness signal of stage ${i+1}$  given step $i$ decreases. This loss in sensitivity explains the decrease in the growth rate of the average fitness of $U^{f_1}$. We call the ``wastage" of fitness queries described here \emph{mutational drag}.

                                                                                                         We conceived of the following mechanism for curbing mutational drag in $U^{f_1}$. This mechanism relies on parameters $\texttt{flagFreq}\in[0, 0.5]$, $\texttt{unflagFreq}\in[\texttt{flagFreq}, 0.5]$, and $\texttt{flagPeriod}$.  If the one-frequency of some locus at the beginning of some generation is less than $\texttt{flagFreq}$, or greater than $1-\texttt{flagFreq}$, then that locus is flagged. Once flagged, a locus remains flagged as long as the one-frequency of the locus is less than $\texttt{unflagFreq}$, or greater than $1-\texttt{unflagFreq}$ at the beginning of each subsequent generation. If a flagged locus in some generation $t$ has remained constantly flagged for the last $\texttt{flagPeriod}$ generations, then the locus is considered to have passed our fixation test, and is not mutated in generation $t$. We call this mechanism \emph{clamping}, because we expect that in the absence of mutation, a locus that has passed our fixation test will quickly go to strict fixation, i.e. the one-frequency of this locus will get ``clamped" at zero or one for the remainder of the run.

\begin{figure*}[tb!]\begin{center}
                                                                                                         \subfigure[Performance of the $U_c^{f_1}$]{\includegraphics[width=7cm]{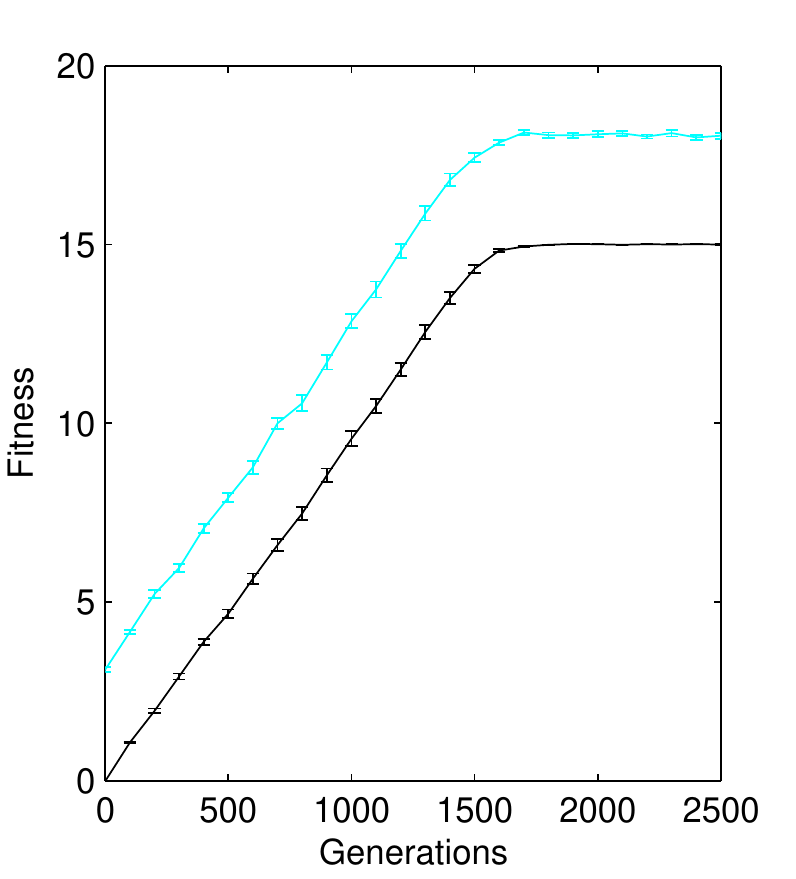}}\quad\quad\quad
                                                                                                         \subfigure[Unmutated Loci in UGA $U_c^{f_1}$]{\includegraphics[width=7cm]{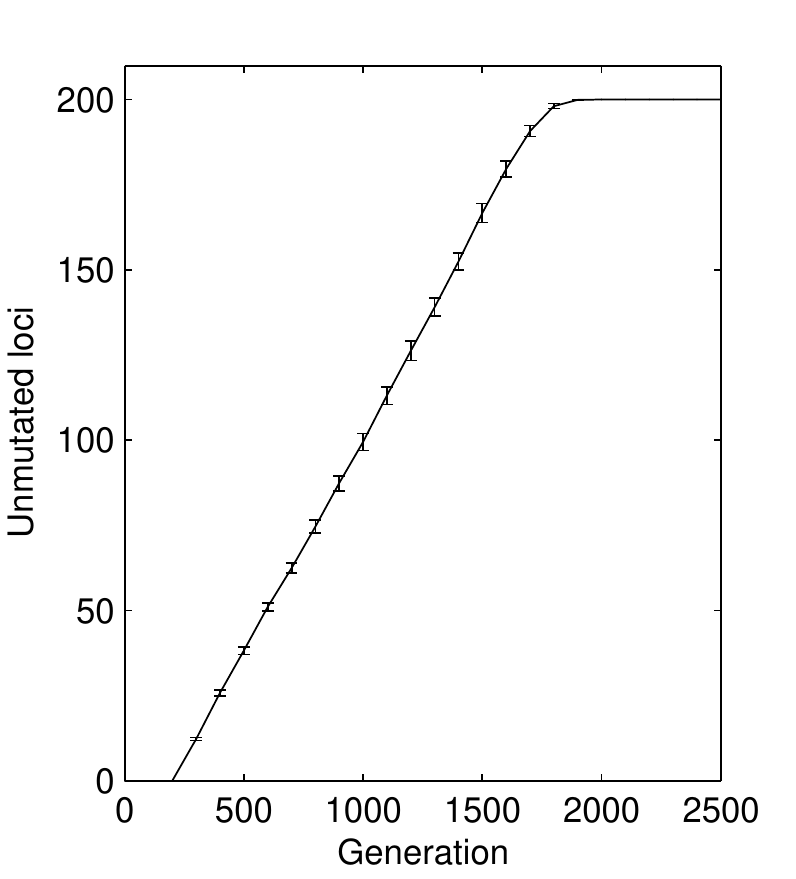}}
                                                                                                         \end{center}
                                                                                                         \caption{\label{staircaseClamping}\emph{(Left:)} The performance, over 20 trials, of the semi-parameterized UGA $U_c$ on the staircase function $f_1$. The mean (across trials) of the average fitness of the population is shown in black. The mean of the best-of-population fitness is shown in blue. \emph{(Right:)} The mean number of loci left unmutated by the clamping mechanism. Errorbars show one standard error above and below the mean every hundredth generation}
                                                                                                         \end{figure*}

                                                                                                         We ran a semi-parameterized UGA $U_c$ which used the clamping mechanism described above and was identical to the semi-parameterized UGA $U$ in every other way on the staircase function $f_1$.  The clamping mechanism used by $U_c$ was parameterized as follows: $\texttt{flagFreq}=0.01$, $\texttt{unflagFreq}=0.1$, $\texttt{flagPeriod=200}$. The performance of $U^{f_1}_c$ is displayed in figure \ref{staircaseClamping}a. Figure \ref{staircaseClamping}b shows the number of loci that the clamping mechanism left unmutated in each generation. These two figures show that the clamping mechanism effectively allowed $U_c$ to climb all the steps of $f_1$. Animation \ref{ctype2StaircaseClampingAnim} shows the one-frequency dynamics of a single run of $U_c^{f_1}$. The action of the clamping mechanism can be seen in the absence of `jitter' in the one-frequencies of loci that have been fixed for a while .

                                                                                                         \subsection{Genoclique Fixing}

                                                                                                         We call a small set of co-adaptive genes an \emph{genoclique}. It is important to stress two features of this definition. Firstly, our use of the term ``co-adaptive", as opposed to the more conventionally used ``co-adapted", is meant to indicate that genocliques are not static entities but dynamic ones that can arise or fade away (become salient, or loose saliency) as the composition of a population of genomes changes. Secondly, note that we have made no commitment to the kind of linkage that must exist between the genes in a genoclique. Linkage between such genes can be weak, or even non-existent.

                                                                                                         Based on the results in the previous sections, we submit that adaptation in simple recombinative genetic algorithms is driven by the recursive fixing of genocliques. We call this the \emph{genoclique fixing hypothesis}.

                                                                                                         This hypothesis rests on assumptions about the distribution of fitness that are easily seen to be \emph{weaker} than those underlying the building block hypothesis \cite{DBLP:journals/corr/abs-0810-3356}---the genoclique fixing hypothesis does not, for example, require large numbers of genes to be individually advantageous at the outset of an evolutionary run. Note, secondly, that genoclique fixing is intuitively a more viable explanation than the building block hypothesis: Because the ability of recombination to disrupt a genoclique declines rapidly as the genoclique goes to fixation, it is easy to see how the fixing of genocliques can be a robust vehicle for adaptation in recombinative genetic systems; in comparison it is much more difficult to grasp how synergistic composition can be a robust vehicle for adaptation. After all, though recombination can occasion the synergistic composition of genes, it can also occasion the destruction of such compositions. Thirdly, note that unlike the building block hypothesis, for which no proof of concept has been provided in over three decades, the genoclique fixing hypothesis is accompanied by proof of concept (see the previous section) from the start.

                                                                                                          \subsection{Empricial Validation}

                                                                                                         We now present the results of an experiment in which the use of clamping dramatically improved the performance of a UGA on large, randomly generated instances of the MAX 3-SAT problem. This difference in performance strongly supports our hypothesis.

\begin{figure*}[tb!]\begin{center}
                                                                                                         \subfigure[Performance of the UGA $Q_c$]{\includegraphics[width=7cm]{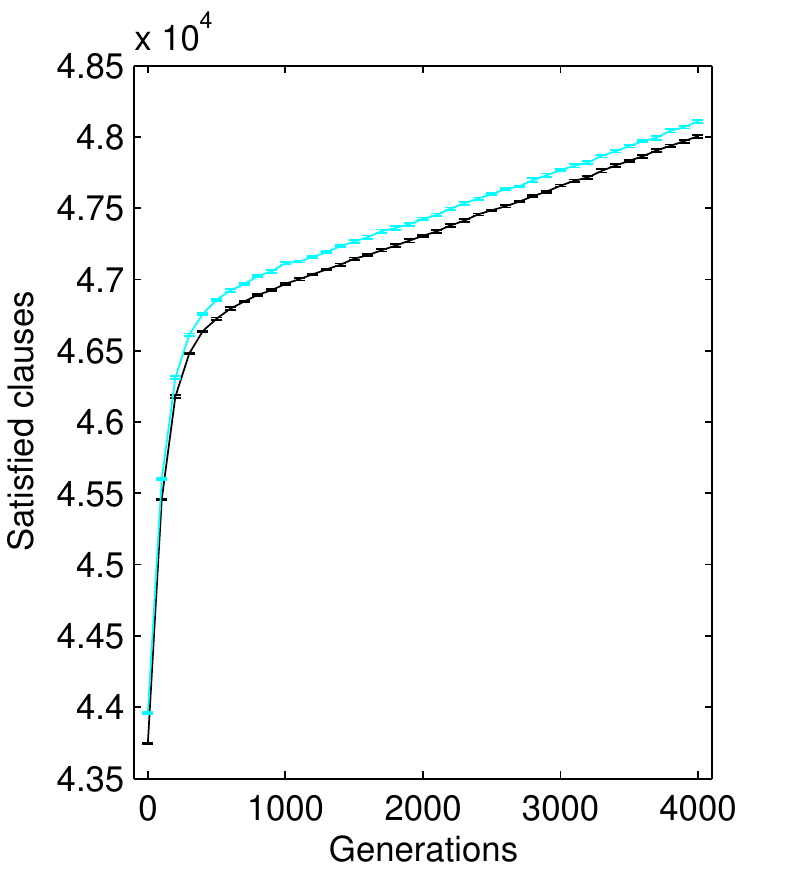}}\quad\quad\quad
                                                                                                         \subfigure[Performance of the UGA $Q$]{\includegraphics[width=7cm]{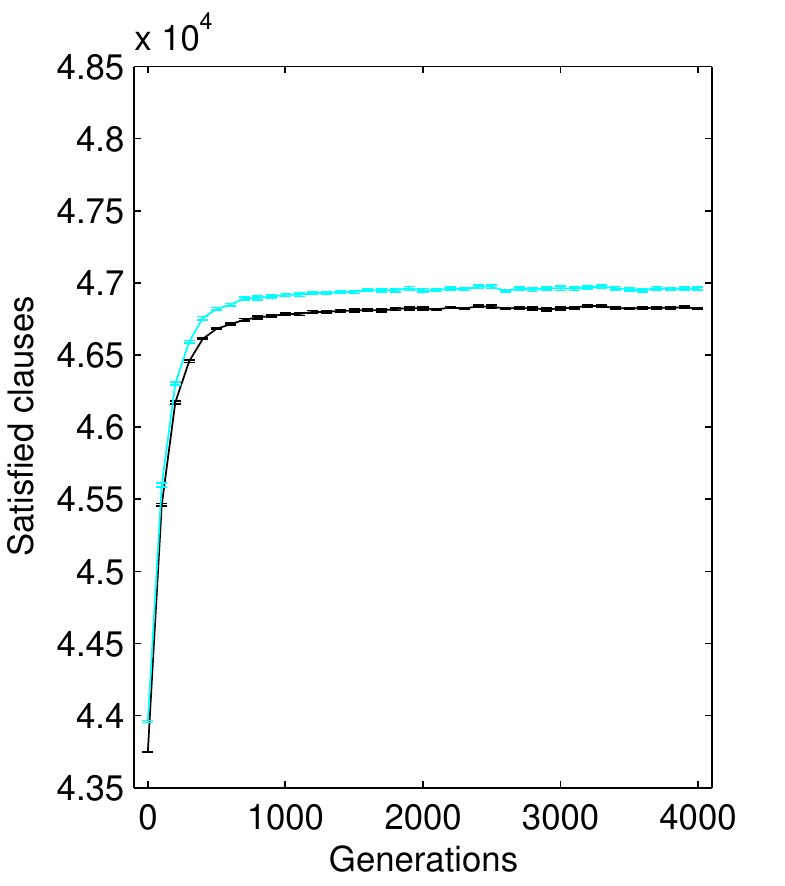}}\\
                                                                                                         \subfigure[Unmutated Loci in UGA $Q_c$] {
                                                                                                         \includegraphics[width=7cm]{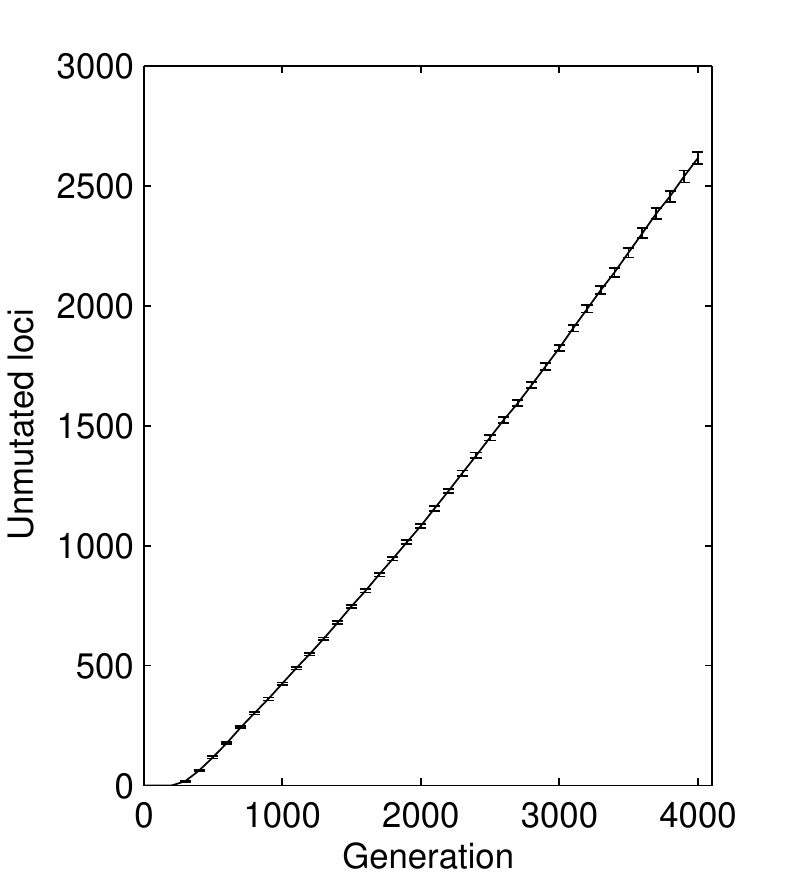}}
                                                                                                         \end{center}
                                                                                                         \caption{\label{performanceMaxsat}\emph{(Top:)} The performance, over 10 trials, of the UGA $Q_c$ (left) and the UGA $Q$ (right), on randomly generated instances of the MAX 3-SAT problem with 10,000 variables and 50,000 clauses. $Q_c$ used clamping, whereas $Q$ did not. The mean (across trials) of the average fitness of the population is shown in black. The mean of the best-of-population fitness is shown in blue. \emph{(Bottom:)} The mean number of loci left unmutated by the clamping mechanism of $Q_c$. Errorbars show one standard error above and below the mean every hundredth generation}
\end{figure*}

                                                                                                         We ran two semi-parameterized UGAs---one with clamping ($Q_c$), and one without $(Q)$---on randomly generated instances of the MAX 3-SAT problem \cite{hoos2004} with 10,000 binary variables and 50,000 clauses. Both UGAs used a straightforward encoding in which each bit of a genome represents the value of a single MAXSAT variable. The fitness of a genome was simply the number of clauses satisfied under the variable assignment represented by the genome. The clamping mechanism used by $Q_c$ was parameterized as follows: $\texttt{flagFreq}=0.01$, $\texttt{unflagFreq}=0.1$, $\texttt{flagPeriod=200}$. Figure \ref{performanceMaxsat}c  shows the number of loci that this mechanism left unmutated in each generation. By the four thousandth generation, the clamping mechanism left on average over 2500 loci unmutated. Given any set of $2500$ loci, in the absence of clamping the chance that the 2500 loci will all go unmutated in some genome is $0.997^{2500}<5.5\times10^{-4}$. The ``drag" resulting from the continued mutation of long-fixed loci in $Q$ explains why this UGA was outperformed by $Q_c$ (Figure \ref{performanceMaxsat}a,b). The difference between the mean best-of-population fitness of the final generation of $Q_c$ and the mean best-of-population fitness of the final generation of $Q$  was 1148.5 clauses. By all indications, this difference would have been larger had we allowed our trials to continue past 4000 generations.

                                                                                                         \section{\label{functionOfRecombination}On the Function of Recombination}

                                                                                                         Under the building block hypothesis, the function of recombination is clear---to drive adaptation by effecting the synergistic composition of advantageous genes, and co-adapted sets of advantageous genes. If genoclique fixing, not synergistic composition, is the vehicle for adaptation, then the function of recombination is less transparent. If the genoclique fixing hypothesis is to be a viable alternative to the building block hypothesis, the advantage that recombination often confers must be accounted for.

                                                                                                         Under the genoclique fixing hypothesis, the widely reported efficacy of recombination, especially strong  forms of recombination, like uniform crossover, actually seems anomalous. As the expected number of crossover points increases, the size of the genes in a genoclique decreases, and the number of genes in a genoclique therefore tends to decrease.  Since genocliques with fewer genes are less likely to be disrupted by recombination, and since the disruption of genocliques \emph{hampers} their fixation, it seems like the fewer the expected number of crossover points in a crossover operation, the better.

                                                                                                         The phenomenon of hitchhiking \cite{Schaffer1991a,forrest93relative} seems to offer an easy explanatory escape from this anomaly. It is simple to see how, as the size of the genes in a genoclique  increases, some situated bit $b$ can become part of one or more genocliques even though it does not contribute to the co-adaptivity of any of these genocliques. If any of the genocliques go to fixation then so will $b$ (i.e. $b$ will \emph{hitchhike} to fixation). Now, suppose it so happens that the complement of $b$ is  implicated in the co-adaptivity of one or more genocliques later on in the evolutionary run. It seems reasonable to suspect that the prior \emph{spurious fixation} of $b$ will prevent any genocliques containing the complement of $b$  from going to fixation.

                                                                                                         Since the prevalence of hitchhiking increases in inverse relation to the expected number of crossover points, it seems plausible that the relative absence of hitchhiking in UGAs can account for the widely reported efficacy of uniform crossover. The prevalence of hitchhiking will be most extreme when recombination is entirely absent. To test our hunch about the utility of recombination, we therefore switched off crossover in the semi-parameterized UGA $U$ and applied the resulting semi-parameterized mutation-only simple genetic algorithm (MGA), denoted  $M$, to the staircase function $f_1$. A comparison between Animations \ref{crosstype2Mut003Anim}, and \ref{crosstype0Mut003Anim} confirms the prevalence of hitchhiking in  $M^{f_1}$ (note how the one-frequencies of high-index loci rush to one or zero at the beginning of the run even though selection is not acting at these loci), and it's relative absence in $U^{f_1}$ (while the one-frequencies of high-index loci do diverge from 0.5, they do so relatively slowly).  Remarkably, despite the prevalence of hitchhiking, $M^{f_1}$ outperforms $U^{f_1}$ (compare Figure \ref{crosstype=0Performance}b with Figure \ref{crosstype=0Performance}a). Figure \ref{crosstype0Frequencies}b, and  Animation \ref{crosstype0Mut003Anim} show that, like $U^{f_1}$, $M^{f_1}$ performs adaptation by implementing hyperclimbing. The difference in performance seen when comparing Figure \ref{crosstype=0Performance}a  with Figure \ref{crosstype=0Performance}b turns out to be representative of a systematic difference in the performance of UGAs and MGAs on basic staircase functions. In an informal empirical comparison of the performance of these SGAs over a broad parametric regime we found that switching off recombination typically \emph{improves} performance.

                                                                                                         The implications of these results for the genoclique fixing hypothesis are mixed. On the one hand, the ``easy explanatory escape" that hitchhiking seemed to offer turns out not to be quite so easy. If anything, the widely observed efficacy of recombination is now \emph{more} puzzling than before.

                                                                                                         On the other hand, the observed hyperclimbing behavior of MGAs on staircase functions reveals the centrality of \emph{fixing} to adaptation in \emph{all} SGAs. To see why, observe that the conclusions we reached by exploiting the symmetries of unparameterized UGAs with staircase fitness functions hold even when uniform crossover is switched off. This realization entails that MGAs, like UGAs, are capable of efficient hyperclimbing\footnote{The building block hypothesis is decidedly silent when it comes to explaining the adaptive capacity of non-recombinative genetic algorithms \cite[p147-155]{Fogel:1995:ECT}. With the discovery that MGAs can implement efficient hyperclimbing, these reports can now be accounted for.}. In terms of the expected number of crossover points per crossover operation, MGAs and UGAs occur at opposite ends of a continuum. Since both these SGAs are capable of efficient hyperclimbing, hyperclimbing seems well positioned to serve as the organizing idea for the study of adaptation in \emph{all} SGAs.

                                                                                                         \subsection{Multi-Staircase Functions}

                                                                                                         Returning to the task of explaining the function of recombination, we conjecture that staircase functions, illuminative as they are, fail to capture some key feature that is commonly present in fitness distributions induced through the representational choices of GA practitioners. We conjecture, furthermore, that hitchhiking interferes with an MGA's ability to exploit this feature.

                                                                                                         Observe that when a UGA is applied to a staircase function, genocliques will tend to become salient \emph{sequentially}. This need not be true when recombinative SGAs are applied to real-world problems. Might hitchhiking pose more of a problem when genocliques become salient \emph{concurrently}? To test this hunch we conceived of the class of \emph{multi-staircase functions}---a straightforward generalization of the class of staircase functions.

                                                                                                          \begin{definition} A multi-staircase function descriptor is a tuple $(c, h, o, \delta, \sigma, \ell, L^{(1)}, \ldots, L^{(c)}, V^{(1)}, \ldots, V^{(c)})$ where  $c, h$,  $o$ and $\ell$ are positive integers with $cho\leq\ell$, $\delta$ and $\sigma$ are positive real numbers, and $L^{(1)},\ldots,L^{(c)}$ and $V^{(1)},\ldots,V^{(c)}$ are matrices with $h$ rows and $o$ columns such that the elements of $L^{(1)}, \ldots,L^{(c)}$ are distinct integers from the set  $[\ell]$ (i.e. $L^{(k_1)}_{i_1j_1}\not =L^{(k_2)}_{i_2j_2}$ unless $i_1=i_2\wedge j_1=j_2 \wedge k_1 =k_2$), each row in each of the matrices $L^{(1)}, \ldots,L^{(c)}$ is sorted in ascending order, and the elements of $V^{(1)}, \ldots,V^{(c)}$ are binary digits.
                                                                                                          \end{definition}

                                                                                                          The function described by a multi-staircase function descriptor $(h, o, \delta,\sigma, \ell, L^{(1)}, \ldots, L^{(c)}, V^{(1)}, \ldots, V^{(c)})$ is the stochastic function over the set of bitstrings of length $\ell$ given by algorithm 1. We call $c$ the \emph{cardinality} of the multi-staircase function, Like we did with staircase functions, we call $h, o, \delta$, $\sigma$, and $\ell$ the \emph{height, order, increment, noisiness} and \emph{span}  respectively.

                                                                                                          \begin{algorithm}[t]
                                                                                                                   \dontprintsemicolon
                                                                                                                   \KwIn{$g$ is a genome of length $\ell$}\;
                                                                                                                   $y=$\mbox{some value drawn from the distribution $\mathcal N(0,\sigma^2)$}\;
                                                                                                                   \For{j=1 to c}{
                                                                                                                        \For{i=1 to h}{
                                                                                                                            \eIf{$\left(g_{L^{(j)}_{i1}}=V^{(j)}_{i1}\right) \wedge \ldots \wedge \left(g_{L^{(j)}_{io}}=V^{(j)}_{io}\right)$}
                                                                                                                            {$y=y+\delta$}
                                                                                                                            {$y=y-(\delta/(2^o-1))$\;
                                                                                                                            \textbf{break}}
                                                                                                                         }
                                                                                                                   }
                                                                                                                   \KwRet{$y$}\;
                                                                                                                   \caption[boo]{A multi-staircase function with descriptor\\ $(c,h,o,\delta,\sigma,\ell,L^{(1)}, \ldots, L^{(c)}, V^{(1)}, \ldots, V^{(c)})$ }
\end{algorithm}

                                                                                                         Our analogy between ladders and staircase functions can be extended to apply to multi-staircase functions. When the cardinality of a multi-staircase function is one, a single staircase is induced; when the cardinality is two or more, multiple ladders are induced. In the latter case, loci belonging to the steps of a particular staircase may be scattered amongst loci belonging to the steps of other ladders. However, since each locus belongs to no more than one staircase, and since the fitness benefits of climbing separate ladders combine additively, each staircase may be climbed independently; in other words, the ``next step" of several ladders can become salient \emph{concurrently}. The ``degree" of concurrency is determined by the cardinality of the multi-staircase function.

                                                                                                         \subsection{Symmetry Analysis}
                                                                                                         Let $f$ be a multi-staircase function with descriptor $(c, h,o,\delta,\sigma, \ell, L^{(1)}, \ldots, L^{(c)},V^{(1)},  \ldots, V^{(c)})$. We say that this function is \emph{basic} if $\ell=cho$, $L^{(k)}_{ij}=ho(k-1)+o(i-1)+j$, i.e. $L^{(k)}$ is the matrix of integers from $(ho)(k-1)+1$ to $hok$ laid out row-wise, and $V$ is a matrix of ones. If $f$ is basic, then the the first five elements of the descriptor of $f$ determines the remaining elements; we therefore write this descriptor as $(c, h,o,\delta,\sigma)$. Given some multi-staircase function $f$ with descriptor $(c, h,o,\delta,\sigma, \ell, L^{(1)}, \ldots, L^{(c)}, V^{(1)},  \ldots, V^{(c)})$, we define the \emph{basic  form} of $f$ to be the basic multi-staircase function $(c, h, o, \delta, \sigma)$.

                                                                                                         Let $f^*$ be some basic staircase function with descriptor $(c, h,o,\delta,\sigma)$, and let $F$ be the set of all staircase functions with basic form $f^*$. Let $W$ be a semi-parameterized UGA or a semi-parameterized MGA. For any staircase function $f\in F$, let $r^{(t)}_f$ be the probability that the average fitness of the population of $W^{f}$ in generation $t$ is $x$. Then by appreciating the symmetries between the unparameterized UGAs $W^{f^*}$ and $W^{f}$ we can deduce the following equalities between probability distributions: for any generation $t$, $r^{(t)}_{f}=r^{(t)}_{f^*}$.  Thus, for any generation $t$, monte-carlo sampling from $r^{(t)}_{f^*}$,  is equivalent to monte-carlo sampling from $r^{(t)}_{f}$.

                                                                                                         \subsection{Performance of a UGA and an MGA on a Multi-Staircase Function}

\begin{figure*}[tb!]\begin{center}
                                                                                                         \subfigure[Performance of the UGA $U^{f_2}$]{\includegraphics[width=7cm]{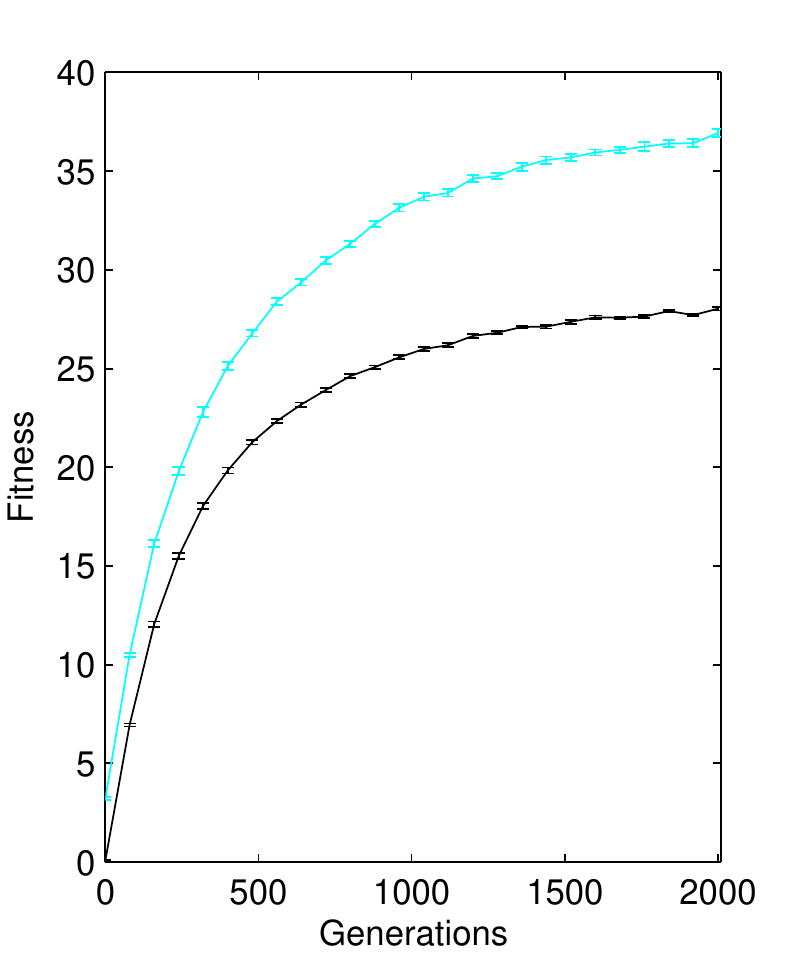}}\quad\quad\quad
                                                                                                         \subfigure[Performance of the MGA $M^{f_2}$]{\includegraphics[width=7cm]{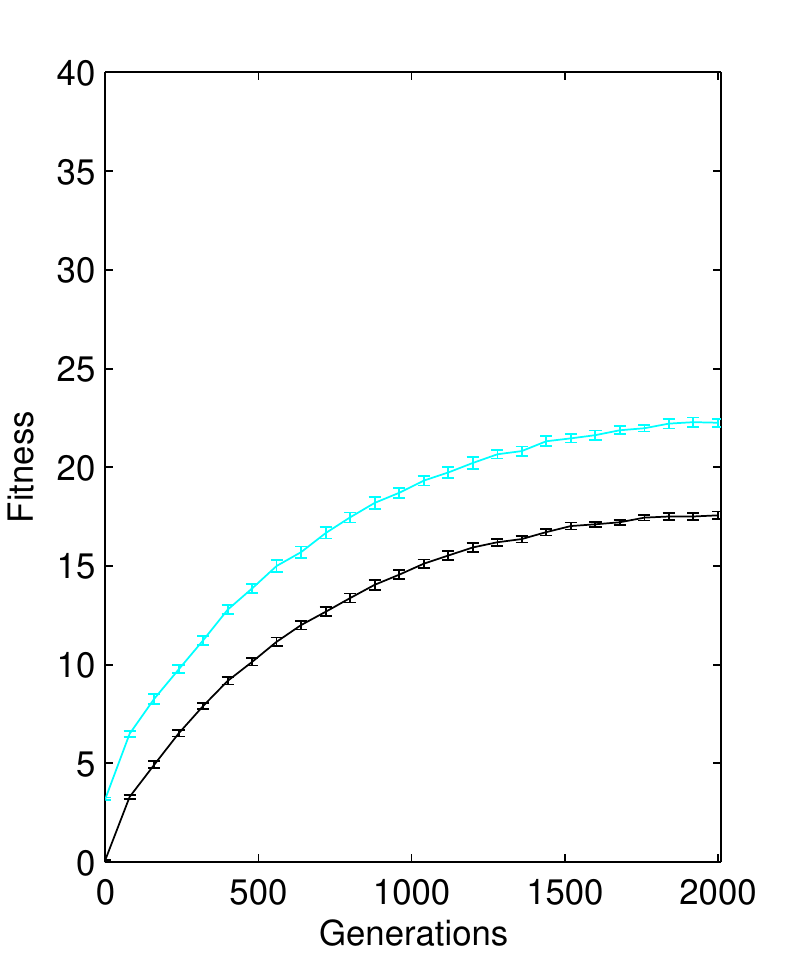}}\end{center}
                                                                                                         \caption{\label{performanceMulti}The performance of the semi-parameterized UGA $U$ (left) and the semi-parameterized MGA $M$ (right) on the multi-staircase function $f_2$ over 20 trials. The mean (across trials) of the average fitness of the population is shown in black. The mean of the best-of-population fitness is shown in blue. The error bars show one standard error above and below the mean every $80^{th}$ generation}
\end{figure*}

                                                                                                         Let $f_2$ denote a multi-staircase fitness function with descriptor $(c=10, h=50, o=4, \delta=0.3, \sigma=1)$. Figure \ref{performanceMulti} shows that when applied to this function, on average the semi-parameterized UGA  $U$ outperforms the semi-parameterized MGA $M$. Animations 3 and 4 show the one-frequency dynamics of  $U^{f_2}$ and $M^{f_2}$ in a single run of each. These animations qualitatively show that $U^{f_2}$ is better than  $M^{f_2}$ at climbing the ten ladders of $f_2$ \emph{in parallel}. The prevalence of hitchhiking in $M^{f_2}$, and it's relative absence in $U^{f_2}$ seems, at least qualitatively, to account for this difference in ability.

                                                                                                         \subsection{Concurrent Genoclique Fixing}
                                                                                                         We emphasize that the semi-parameterized SGAs $U$ and $M$ mentioned above are the same semi-parameterized SGAs that were used in our previous experiments. Recall that on average $M$ outperformed $U$ when applied to the basic staircase function $f_1$. This function can be thought of as a basic \emph{multi}-staircase function with cardinality one. When $f_1$ is regarded as such, the difference between it and $f_2$ amounts solely to a difference in cardinality. Based on these observations, and the results mentioned above, we submit that the function of recombination in genetic algorithms is to reduce hitchhiking; by reducing hitchhiking, recombination allows the fixing of genocliques to proceed concurrently.

                                                                                                         \section{Conclusion}

                                                                                                         Many details of the new theory presented in this paper remain to be worked out and/or expressed. For example, the function of mutation needs to be explained (if mutation causes drag, why use it?), and the relationship between population size and a recombinative SGA's capacity for efficient genoclique fixing merits attention. Presenting a complete account of the workings of recombinative SGAs, however, is not our aim. Rather, we have sought to present a \emph{general} account of these workings, and to support this account in ways that make it compelling, or, to be precise, more compelling  than the building block hypothesis---to date, the only other general account of the practical workings of recombinative SGAs.

                                                                                                         Perhaps the best way to understand the difference between the building block hypothesis and the genoclique fixing hypothesis is by focusing on the part played by fixation in each account. In downplaying the role of fixation, the building block hypothesis departs rather radically from the accounts about adaptation in biological populations that one finds in population genetics. The building block hypothesis holds that genetic algorithms work by maintaining a store of partial solutions---advantageous genes, and coadapted sets of individually advantageous genes---and by hierarchically assembling these partial solutions as evolution proceeds. Crucially, the building block hypothesis is not opposed to the idea that an advantageous gene and it's advantageous bitwise complement can both persist in an evolving population. Indeed, as Watson's work with hierarchical if and only if functions \cite{oai:eprints.ecs.soton.ac.uk:12006,watsonBook} indicates, the persistence of such alleles is \emph{expected}. Because the building block hypothesis dispenses with fixation, it needs to look to the weakness of recombination as a vehicle for "locking in" adaptive gains. This hypothesis cannot, therefore, explain the widely observed adaptive capacity of SGAs with strong forms of recombination (e.g. uniform crossover)

                                                                                                         In contrast, the genoclique fixing hypothesis holds that \emph{fixation} is the vehicle by which adaptive gains are locked in. The genoclique fixing hypothesis is based on the key realization that selection can drive a small set of unlinked coadapted genes to fixation even as these genes are repeatedly separated by recombination whenever they co-occur \cite{DBLP:journals/corr/abs-0810-3357}. Once such a  set of genes---what we call a \emph{genoclique}---has gone to fixation,  recombination looses it's power to disrupt this set, and the fitness advantage that the genoclique confers, even if it is only a small increase in expected fitness, gets locked in. Since recombination is not required to ``protect" genocliques as they go to fixation, the genoclique fixing hypothesis has no problem in accounting for the adaptive capacity of UGAs. So, while the building block hypothesis can only account for the adaptive capacity of SGAs with small numbers of crossover points, the genoclique fixing hypothesis can account for the adaptive capacity of \emph{any} recombinative SGA.

                                                                                                          The genoclique fixing hypothesis can be thought of as a particular instantiation of a more general \emph{unified} theory about the practical workings of  \emph{all} SGAs, including ones that do not use uses crossover. In section \ref{hyperclimbing} we introduced the idea of a hyperclimbing heuristic. This heuristic is sensitive, not to the local features of a search space, but to fitness properties of the hyperplanes of the space. The hyperclimbing heuristic is therefore not susceptible to the typical problems affecting local search algorithms (e.g. entrapment in the fitness basins of local optima). While hyperclimbing seems like a reasonable way to perform adaptive search, the moment one factors in what appears to be the high cost, in terms of time and fitness queries, of implementing this heuristic, it quickly looses it's shine.  Our exciting discovery---the crux of this paper---is that simple genetic algorithms can implement hyperclimbing efficiently.

                                                                                                         On the problems studied, we found that an SGA with uniform crossover, and an SGA without crossover can both perform efficient hyperclimbing. Uniform crossover and no crossover are, in terms of expected number of crossover points, at opposite ends of the ``crossover continuum" of an SGA. We therefore infer that a capacity for efficient hyperclimbing underlies the adaptive capacity of all SGAs. We submit this idea---the \emph{hyperclimbing thesis}---as a platform for the unified study of adaptation in all genetic algorithms.

                                                                                                       \bibliographystyle{plain}
                                                                                                       \bibliography{c:/mystuff/mycreations/0Work/refs}
                                                                                                       \pagebreak
                                                                                                       \begin{appendix}
                                                                                                       \section*{Materials and Methods}
                                                                                                         The semi-parameterized SGA denoted  by $U$ was implemented with an SGA that is faithful to the specification for a simple genetic algorithm given by Mitchell \cite[p 10]{Mitchell:1996:IGA} in every way, except for the following two:
                                                                                                         \begin{enumerate}
                                                                                                         \item In each generation, right after evaluating the fitness of all individuals, our SGA used sigma scaling \cite[p 167]{Mitchell:1996:IGA} to adjust the fitness of each individual, and used this adjusted fitness when selecting the parents of that generation. Suppose $f^{(t)}_x$ is the fitness of some individual $x$ in some generation $t$, and suppose the average fitness and standard deviation of the fitness of the individuals in generation $t$ are given by $\overline{f^{(t)}}$ and  $\sigma^{(t)}$ respectively, then the adjusted fitness of $x$ in generation $t$ is given by $h^{(t)}_x$ where, if $\sigma^{(t)}=0$ then $h^{(t)}_x=1$, otherwise,  $$h^{(t)}_x=\min(0,1+\frac{f^{(t)}_x -\overline{f^{(t)}}}{\sigma^{(t)}})$$
                                                                                                         \item The SGA used universal stochastic stochastic sampling \cite{Baker:1985:ASM} \cite[p 166]{Mitchell:1996:IGA} to select parents.
                                                                                                         \end{enumerate}

                                                                                                         Selection was fitness-proportionate. The population size was 500. Bit-flip mutation with a mutation rate of $0.003$ per bit was used. The probability of crossover was one.

                                                                                                         The population size of the semi-parameterized UGAs $Q_c$ and $Q$ was 200. $Q_c$ used clamping (described in the main text), whereas $Q$ did not. Other than the population size, and the use of clamping, $Q_c$ and $Q$ were the same in every way to the semi-parameterized UGA $U$. The SGA used to implement the semi-parameterized SGAs described above was written in Matlab and is available for download\footnote{The SGA and all fitness functions used in this paper can be downloaded
	from \url{http://www.cs.brandeis.edu/~kekib/GAWorkingsMatlab.zip}}.

                                                                                                       \section*{Proofs}
                                                                                                       \begin{lemma}For any staircase function with descriptor $(h,o,\delta,\sigma,\ell, L,V)$, and any integer $i\in[h]$, the fitness signal of step $i$ is $i\delta $.
                                                                                                       \end{lemma}
                                                                                                       \noindent \textsc{Proof:} The proof is by induction on  $i$. The base case, when $i=h$ is easily seen to be true. For any $k\in\{2,\ldots,h\}$, we assume that the hypothesis holds for $i=k$, and prove that it holds for $i=k-1$.  For any $j\in[h]$, let $\gamma_j$ denote stage $j$, and let $\Gamma_j$ be the canonical denotation of the schema partition containing $\gamma_j$. The fitness signal of step $k-1$ is given by \[\frac{1}{2^o}\left(S_{\Gamma_1\ldots\Gamma_k}(\gamma_1\ldots \gamma_k)+\sum_{\psi\in\Gamma_k\backslash\{\gamma_k\}}S_{\Gamma_1\ldots\Gamma_k}(\gamma_1\ldots\gamma_{k-1}\psi)\right)\]
                                                                                                       \[=\frac{\delta k}{2^o}+\frac{2^o-1}{2^o}\left(\delta(k-1)-\frac{\delta}{2^o-1}\right)\]
                                                                                                       where the first term of the right hand side follows from the inductive hypothesis.
                                                                                                       Manipulation of the right hand side yields \[\frac{\delta k+(2^o-1)\delta(k-1)-\delta }{2^o}\] which upon further manipulation yields $(k-1)\delta $ $\Box$

                                                                                                       \begin{corollary} For any $i\in\{2,\ldots,h\}$, the conditional signal to noise ratio of stage $i$ given step $i-1$ is $\delta/\sigma$
                                                                                                       \end{corollary}
                                                                                                       \textsc{Proof} The conditional signal to noise ratio of stage $i$ given step $i-1$ is given by
                                                                                                       \begin{align*}
                                                                                                       &S_{\Gamma_i|\Gamma_1\ldots\Gamma_{i-1}}(\gamma_i|\gamma_1\ldots\gamma_{i-1})/\sigma\\
                                                                                                       &=(S_{\Gamma_1\ldots\Gamma_i}(\gamma_1\ldots\gamma_i)-S_{\Gamma_1\ldots\Gamma_{i-1}}(\gamma_1\ldots\gamma_{i-1}))/\sigma\\
                                                                                                       &=(i\delta -(i-1)\delta)/\sigma\\
                                                                                                       &=\delta/\sigma\,\,\Box \end{align*}
                                                                                                       \begin{theorem} For any staircase function with descriptor $(h,o,\delta,\sigma,\ell, L,V)$, and any integer $i\in[h]$, the fitness signal of stage $i$ is $\delta/(2^o) ^{i-1}$.
                                                                                                       \end{theorem}
                                                                                                       \noindent \textsc{Proof:} For any $j\in[h]$, let $\gamma_j$ denote stage $j$, and let $\Gamma_j$ be the canonical denotation of the partition containing $\gamma_j$.
                                                                                                       We first prove the following claim
                                                                                                       \begin{claim} For any $i\in[h]$, \[\sum_{\xi_1\in\Gamma_1\ldots\Gamma_i\backslash\{\gamma_1\ldots\gamma_i\}}S_{\Gamma_1\ldots\Gamma_i}(\xi)=-i\delta\]\end{claim}
                                                                                                       The proof of the claim follows by induction on $i$. The proof for the base case $(i=1)$ is as follows:\balance
                                                                                                       \[\sum_{\xi\in\Gamma_1\backslash\{\gamma_1\}}S_{\Gamma_1}(\xi)=(2^o-1)\left(\frac{-\delta}{2^o-1}\right)=-\delta\]
                                                                                                       For any $k\in[h-1]$ we assume that the hypothesis holds for $i=k$ and prove that it holds for $i=k+1$.
                                                                                                       \[\sum_{\xi_1\in\Gamma_1\ldots\Gamma_i\backslash\{\gamma_1\ldots\gamma_{k+1}\}}S_{\Gamma_1\ldots\Gamma_{k+1}}(\xi)\]
                                                                                                       \[\phantom{aaaa}=\sum_{\psi\in\Gamma_{k+1}\backslash\{\gamma_{k+1}\}}S_{\Gamma_1\ldots\Gamma_k+1}(\gamma_1\ldots\gamma_k\psi)+
                                                                                                       \sum_{\xi\in\Gamma_1\ldots\Gamma_k\backslash\{\gamma_1\ldots\gamma_k\}}\,\,\sum_{\psi\in\Gamma_{k+1}}S_{\Gamma_1\ldots\Gamma_{k+1}}(\xi\psi)
                                                                                                       \]
                                                                                                       \[\phantom{aaaa}=\sum_{\psi\in\Gamma_{k+1}\backslash\{\gamma_{k+1}\}}S_{\Gamma_1\ldots\Gamma_k+1}(\gamma_1\ldots\gamma_k\psi)+
                                                                                                       \sum_{\psi\in\Gamma_{k+1}}\,\,\,\,\sum_{\xi\in\Gamma_1\ldots\Gamma_k\backslash\{\gamma_1\ldots\gamma_k\}}S_{\Gamma_1\ldots\Gamma_{k+1}}(\xi\psi)
                                                                                                       \]
                                                                                                       \[\phantom{aaaa}=(2^o-1)\,S_{\Gamma_1\ldots\Gamma_k+1}(\gamma_1\ldots\gamma_k)+
                                                                                                       2^o\left(\sum_{\xi\in\Gamma_1\ldots\Gamma_k\backslash\{\gamma_1\ldots\gamma_k\}}S_{\Gamma_1\ldots\Gamma_{k+1}}(\xi)\right)
                                                                                                       \]
                                                                                                       where the last equality follows from the definition of a staircase function. Using Lemma 1 and the inductive hypothesis, the right hand side of this expression can be seen to equal \[(2^o-1)\left(\delta k-\frac{\delta}{2^o-1}\right)-2^o\delta k\]
                                                                                                       which upon some simple manipulation yields $-\delta(k+1)$.

                                                                                                       For a proof of the theorem, observe that stage 1 and step 1 are the same schema. So, by Lemma 1,
                                                                                                       $S_{\Gamma_{1}}(\gamma_{1})=\delta$. Thus the theorem holds for $i=1$. For any $i\in\{2,\ldots,h\}$,
                                                                                                       \[S_{\Gamma_{k}}(\gamma_{i})=\frac{1}{(2^o)^{i-1}}\left(S_{\Gamma_1\ldots\Gamma_i}(\gamma_1\ldots\gamma_{i})+\sum_{\xi\in\Gamma_1\ldots\Gamma_{i-1}\backslash\{\gamma_1\ldots\gamma_{i-1}\}}S_{\Gamma_1\ldots\Gamma_i}(\xi\gamma_k)\right)\]
                                                                                                       \[\phantom{S_{\Gamma_{k}}(\gamma_{i})}=\frac{1}{(2^o)^{i-1}}\left(S_{\Gamma_1\ldots\Gamma_i}(\gamma_1\ldots\gamma_{i})+\sum_{\xi\in\Gamma_1\ldots\Gamma_{i-1}\backslash\{\gamma_1\ldots\gamma_{i-1}\}}S_{\Gamma_1\ldots\Gamma_{i-1}}(\xi)\right)\]
                                                                                                       where the last equality follows from the definition of a staircase function. Using Lemma 1 and Claim 1, the right hand side of this equality can be seen to equal
                                                                                                       \[\frac{i\delta  -(i-1)\delta}{(2^o)^{i-1}}\]
                                                                                                       \[=\frac{\delta}{(2^o)^{i-1}}\quad\Box\\\]


                                                                                                       \begin{Animation*}[p!]\begin{center}
                                                                                                            \includemovie[ rate=0.7,
                                                                                                            text={\includegraphics{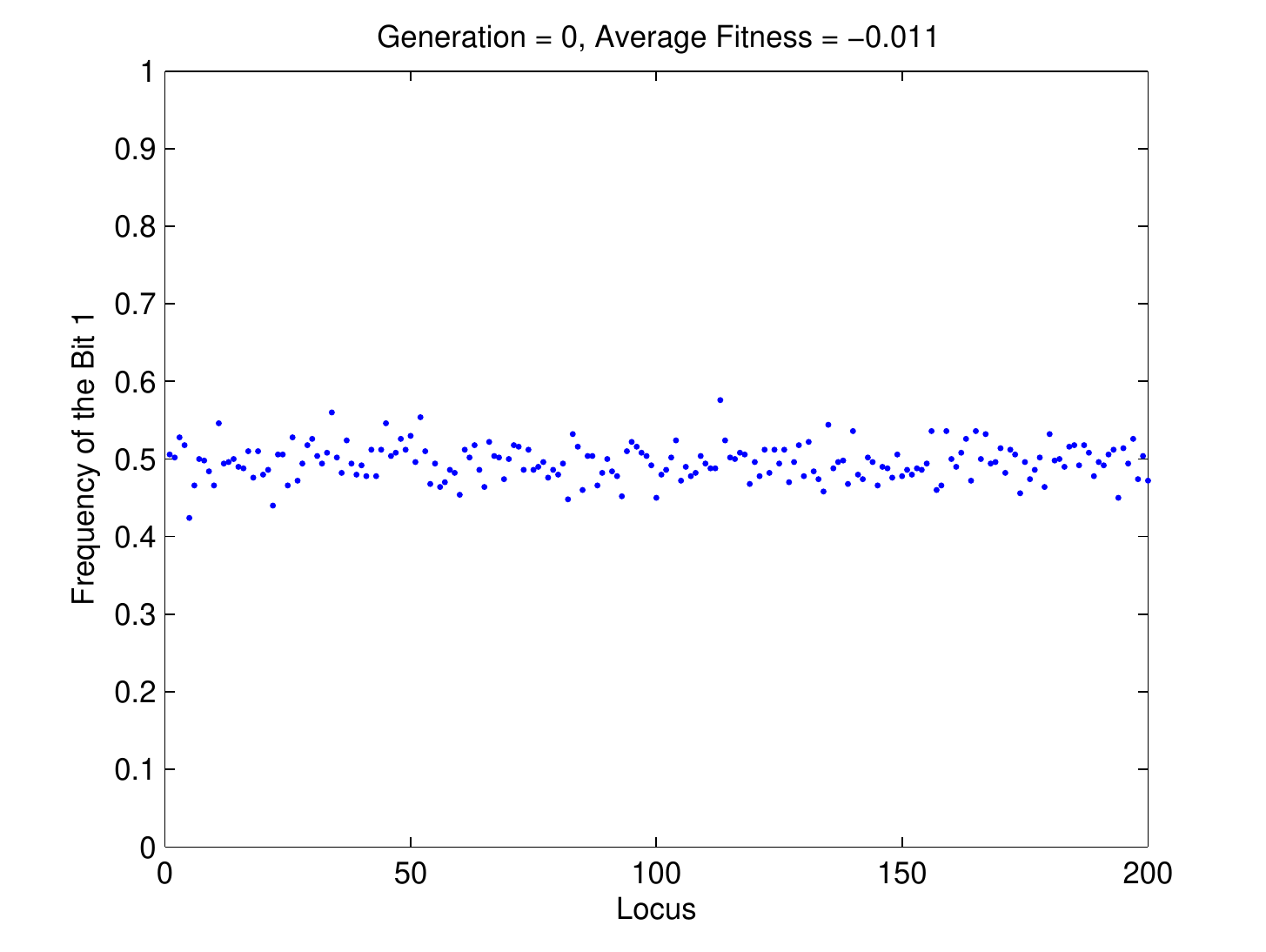}}
                                                                                                            ]{15cm}{11cm}{crosstype=2Mut=003Anim.mpg}\end{center}
                                                                                                            \caption{\label{crosstype2Mut003Anim}[Click on image to play] The one-frequency dynamics of each locus of the UGA $U^{f_1}$ over the first 500 generations of a single run . (If the animation does not work please download the full version of this manuscript from \url{www.cs.brandeis.edu/~kekib/GAWorkings.html})}
                                                                                                       \end{Animation*}

                                                                                                        \begin{Animation*}[p!]\begin{center}
                                                                                                            \includemovie[ rate=0.7,
                                                                                                            text={\includegraphics{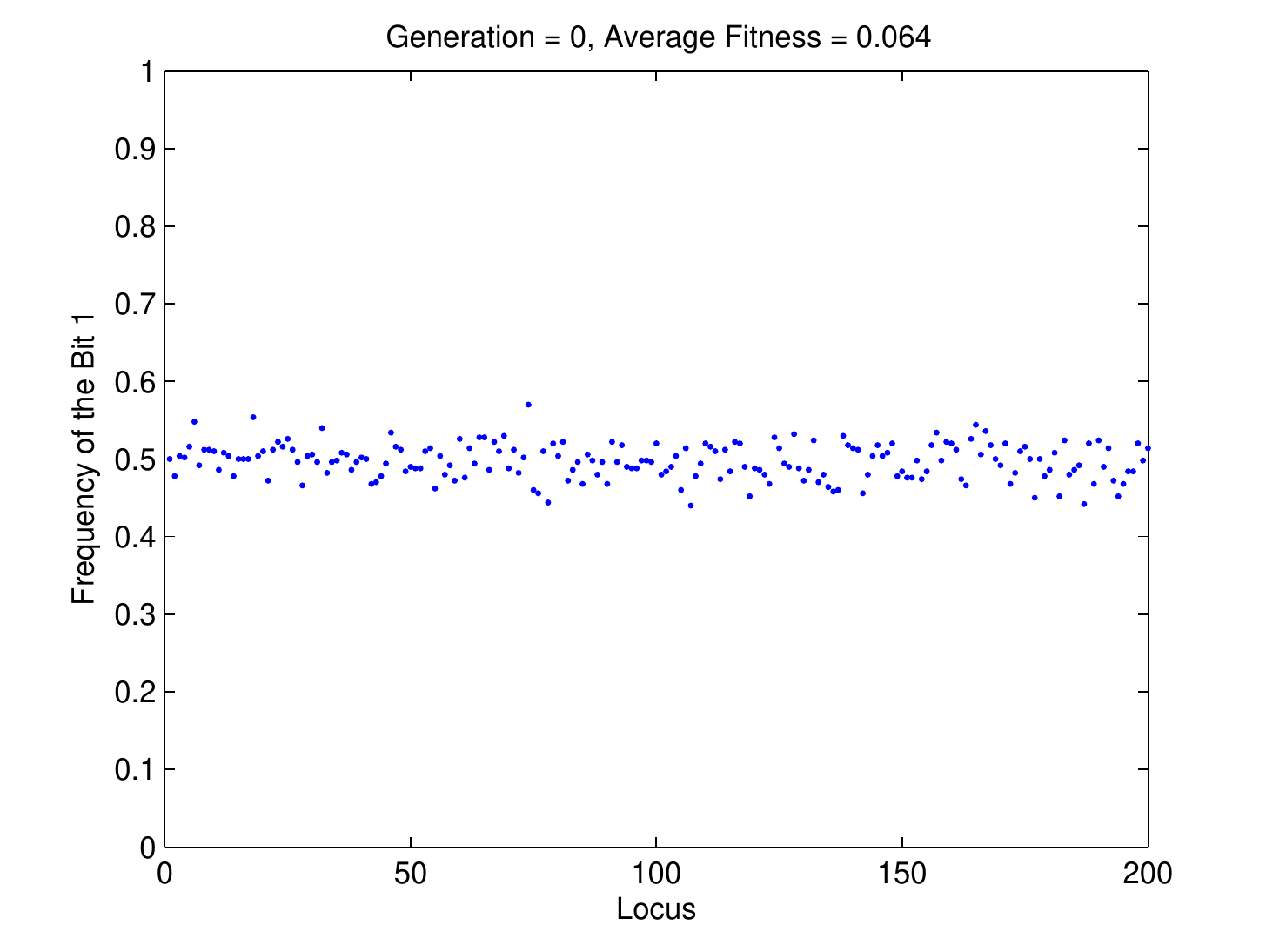}}
                                                                                                            ]{}{}{ctype=2StaircaseClampingAnim.mpg}\end{center}
                                                                                                            \caption{\label{ctype2StaircaseClampingAnim}[Click on image to play] The one-frequency dynamics of each locus of the UGA $U_c^{f_1}$ over the first 500 generations of a single run. (If the animation does not work please download the full version of this manuscript from \url{www.cs.brandeis.edu/~kekib/GAWorkings.html})}
                                                                                                       \end{Animation*}

                                                                                                         \begin{Animation*}[p]\begin{center}
                                                                                                            \includemovie[ rate=0.7,
                                                                                                            text={\includegraphics{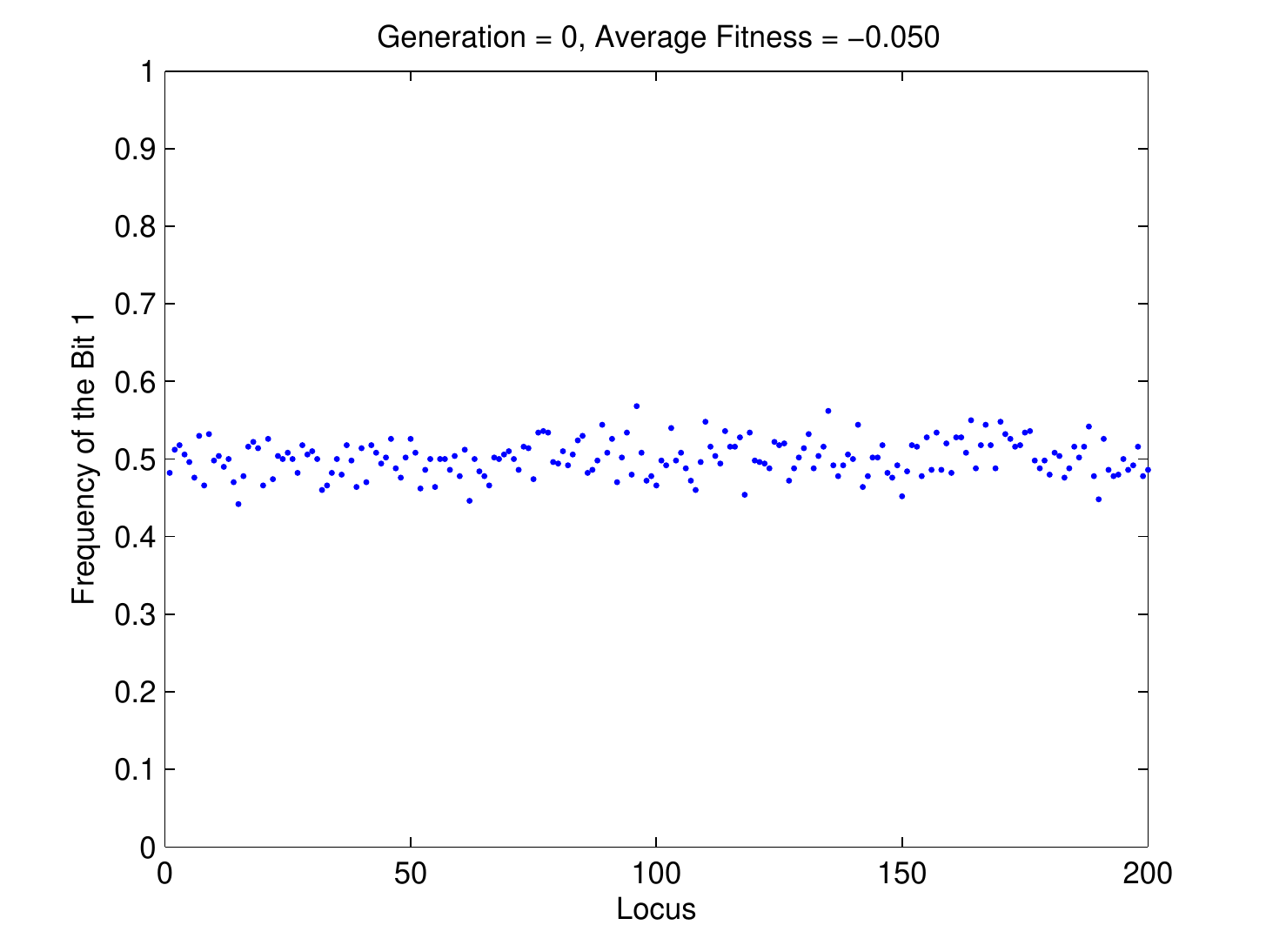}}
                                                                                                            ]{15cm}{11cm}{crosstype=0Mut=003Anim.mpg}\end{center}
                                                                                                            \caption{\label{crosstype0Mut003Anim} [Click on image to play] A visualization of the one-frequency dynamics of each locus over the first 500 generations of a single run of  the MGA $M^{f_1}$. (If the animation does not work please download the full version of this manuscript from \url{www.cs.brandeis.edu/~kekib/GAWorkings.html})}
                                                                                                         \end{Animation*}

                                                                                                       \begin{Animation*}[p]\begin{center}
                                                                                                            \includemovie[ rate=0.7,
                                                                                                            text={\includegraphics{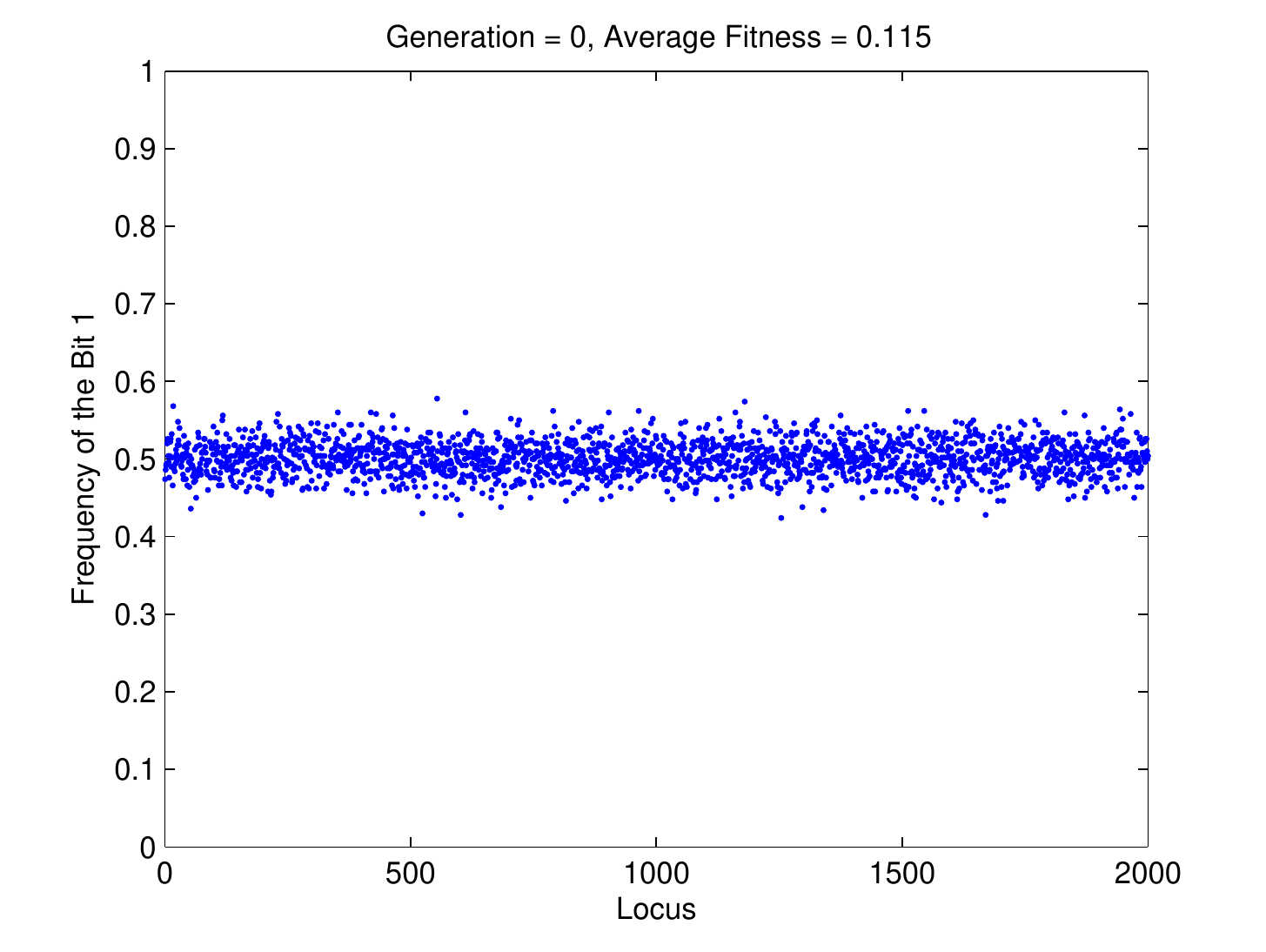}}
                                                                                                            ]{15cm}{11cm}{crosstype=2Mut=003MultiAnim.mpg}\end{center}
                                                                                                            \caption{\label{crosstype2Mut003MultiAnim}[Click on image to play] A visualization of the one-frequency dynamics of each locus over the first 500 generations of a single run of the UGA $U^{f_2}$. (If the animation does not work please download the full version of this manuscript from \url{www.cs.brandeis.edu/~kekib/GAWorkings.html})}
                                                                                                         \end{Animation*}

                                                                                                         \begin{Animation*}[p]\begin{center}
                                                                                                            \includemovie[ rate=0.7,
                                                                                                            text={\includegraphics{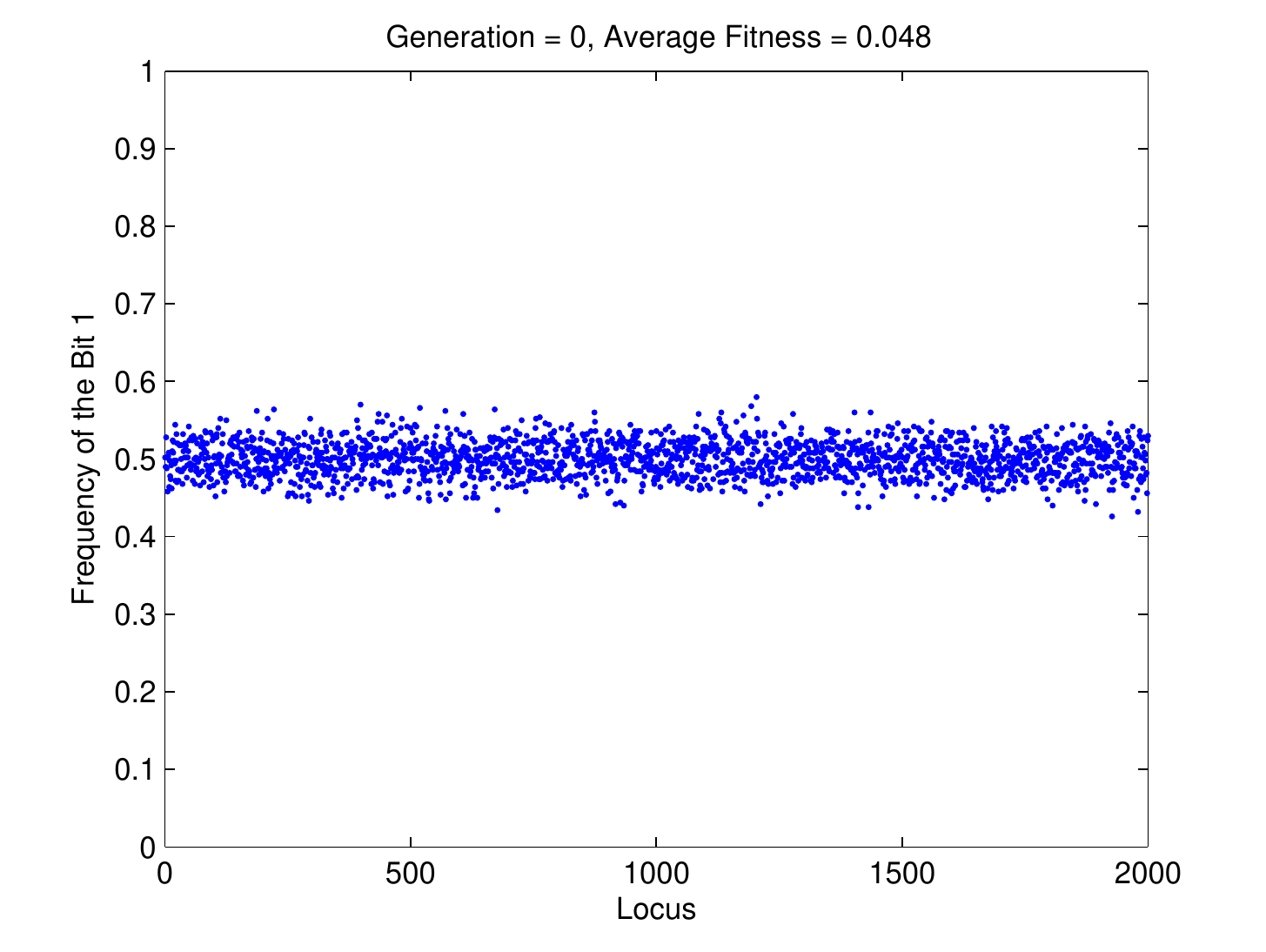}}
                                                                                                            ]{15cm}{11cm}{crosstype=0Mut=003MultiAnim.mpg}\end{center}
                                                                                                            \caption{\label{crosstype0Mut003MultiAnim} [Click on image to play] A visualization of the one-frequency dynamics of each locus over the first 500 generations of a single run of  the MGA $M^{f_2}$. (If the animation does not work please download the full version of this manuscript from \url{www.cs.brandeis.edu/~kekib/GAWorkings.html})}
                                                                                                         \end{Animation*}

                                                                                                      \end{appendix}


                                                                                                        \end{document}